%% file: arXiv_FH_VF_Bound_on_Continuous_Function_Approximation.tex
\documentclass{article}

\usepackage{arxiv}

\usepackage[utf8]{inputenc} % allow utf-8 input
\usepackage[T1]{fontenc}    % use 8-bit T1 fonts
\usepackage{hyperref}       % hyperlinks
\usepackage{url}            % simple URL typesetting
\usepackage{booktabs}       % professional-quality tables
\usepackage{amsfonts}       % blackboard math symbols
\usepackage{nicefrac}       % compact symbols for 1/2, etc.
\usepackage{microtype}      % microtypography
\usepackage{lipsum}		% Can be removed after putting your text content
\usepackage{graphicx}
\usepackage{natbib}
\usepackage{doi}

\title{A General Constructive Upper Bound on Shallow Neural Nets Complexity}

\date{} 					% Or removing it

\author{\href{https://orcid.org/0000-0001-6291-2767}{\includegraphics[scale=0.06]{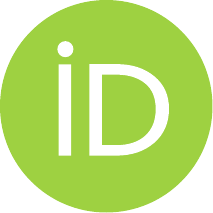}\hspace{1mm}Franti{\v{s}ek Hakl}
        \thanks{Corresponding author.}} \\
	Institute of Computer Science\\
	Department of Artificial Intelligence\\
	Pod Vod\'arenskou v\v{e}\v{z}\'{\i} 2\\
        Prague, Czech Republic \\
	\texttt{hakl@cs.cas.cz} \\
	\And
	V\'{\i}t Fojt\'{\i}k \\
	Ludwig-Maximilians-Universität München\\
	Department of Mathematics\\
	Geschwister-Scholl-Platz 1\\
        München, Germany\\
	\texttt{fojtik@math.lmu.de} \\
}

\renewcommand{\shorttitle}{A Constructive Upper Bound on Neural Nets}

\hypersetup{
pdftitle={A Constructive Upper Bound on Neural Nets},
pdfsubject={q-bio.NC, q-bio.QM},
pdfauthor={Franti{\v{s}ek Hakl, V\'{\i}t Fojt\'{\i}k}},
pdfkeywords={approximation of functions, shallow neural networks, upper estimate of size, Stone-Weierstrass theorem},
}

\usepackage{amssymb}	    % subsetneq (subset but not equal)
\usepackage{amsmath}        % extensions for typesetting of math
\usepackage{amsfonts}       % math fonts
\usepackage{amsthm}         % theorems, definitions, etc.

\usepackage[utf8]{inputenc}

\usepackage{ifthen}
\usepackage{eqnarray}

\usepackage{graphicx,type1cm,eso-pic,color} % needed by watermark !!!
\definecolor{lateralwatermarkred}{rgb}{1.0, 0.3, 0.3}
\definecolor{bodywatermarkred}{rgb}{1.0, 0.50, 0.50}

\definecolor{lateralwatermarkred}{rgb}{1.0, 1.0, 1.0}
\definecolor{bodywatermarkred}{rgb}{1.0, 1.0, 1.0}

\makeatletter
\AddToShipoutPicture{%
\setlength{\@tempdimb}{.5\paperwidth}%
\setlength{\@tempdimc}{.5\paperheight}%
\setlength{\unitlength}{1pt}%
\put(\strip@pt\@tempdimb,\strip@pt\@tempdimc){%
\makebox(-485,-5){\rotatebox{90}{\textcolor{lateralwatermarkred}%
{\fontsize{10mm}{10mm}\selectfont{\textcopyright \hspace{0.4ex} F. Hakl,
    V. Fojtík, Prague, \today -- DRAFT version}}}
}% 
\makebox(0,0){\rotatebox{60}{\textcolor{bodywatermarkred}%
{\fontsize{27mm}{27mm}\selectfont{{\bf DRAFT VERSION}}}}
}%
\makebox(485,-5){\rotatebox{90}{\textcolor{lateralwatermarkred}%
{\fontsize{10mm}{10mm}\selectfont{\textcopyright \hspace{0.4ex} F. Hakl,
     V. Fojtík, Prague, \today -- DRAFT version}}}
}% 
}
}
\makeatother

\newcounter{PrintAllSymbols}  
\input{./NN_common_math_symbols}

\begin{document}
\maketitle

\begin{abstract}
  We provide an upper bound on the number of neurons of a shallow neural
  network required to approximate a function continuous on a compact set with
  given accuracy.  This method, inspired by a specific proof of the
  Stone-Weierstrass Theorem, is constructive and more general than previous
  bounds of this character, as it is applicable to any continuous function on
  any compact set. 
\end{abstract}


\section{Introduction, Results}

One of the more studied theoretical topics in neural network research is the
expressive power of shallow (one-hidden-layer) networks, where several
versions of approximating universality have been proved and widely
studied. Much work focuses on determining lower and upper estimates for the
size of neural networks needed to approximate a given function. 
However, no bounds found in the literature are general and explicit enough to
allow us, in practice, given a continuous function, to estimate the sufficient
and necessary size of a shallow network required to approximate the function
with given accuracy. Furthermore, we are interested in the structure of the
approximating function that satisfies the given estimate. We can obtain this
structure by constructing the given approximating function using knowledge of
the approximated function. The primary objective of this paper is to take a
step in this direction.

Given a compact set $\anset{K}\subset\Real{d}$, a continuous function \funct
f\maps\anset{K}\to{\Real{}} and $\varepsilon > 0$, we construct an upper bound
on the least number $h$ such that there exists a mapping \mapping{g} of a
neural network that has a single hidden layer consisting of $h$ neurons with
exponential transfer function and that satisfies
$\norm{ \mapping{f} - \mapping{g}}{\infty} < \varepsilon$. Our bound depends
on the input dimension $d \geq 2$, on the diameter of the set \anset{K}
denoted \diameter{\anset{K}}\nolinebreak, on the sup norm
\norm{\mapping{f}}{\infty}, and of course on the desired approximation
accuracy $\varepsilon > 0$. Furthermore, the complexity of the target function
\mapping{f} is expressed by the inverse modulus of continuity:

\begin{definition}
	\label{def:invmodulus}
	Let \funct f\maps \anset{K} \to{\Real{}},  $\varepsilon > 0$ and 
	\begin{equation*}
          \anset{{\cal Y}_\delta} \defeq \defset{\left(\vektor{y}_1,\vektor{y}_2\right)\in \karttwoprod{\anset{K}}{\anset{K}}}
          {\HOLDS{\norm{\vektor{y}_1-\vektor{y}_2}{} < \delta}} , 
\end{equation*}

\begin{equation*}
          \anset{{\cal F}_\varepsilon} \!\defeq \! \defset{\!\left(\vektor{y}_1,\vektor{y}_2\right)\in \karttwoprod{\anset{K}}{\anset{K}}}
          {\HOLDS{\absvalue{\val f\in{\vektor{y}_1} - \val f\in{\vektor{y}_2}}
              \leq \varepsilon}\!} \! . 
\end{equation*}
Then define the inverse modulus of continuity of \mapping{f} at $\varepsilon$ as
\begin{equation*}
     \omega^{-1}(\mapping{f}, \varepsilon) \defeq \sup \defset{\delta}
     {\HOLDS{ \anset{\cal Y}_\delta \subset \anset{{\cal F}_\varepsilon} }} \; .
 \end{equation*}
In words, the inverse modulus of continuity of the function
        \mapping{f} for the value $\varepsilon$ is the largest $\delta$ such that if
        the distance between $\vektor{y}_1$ and $\vektor{y}_2$ is less than
        $\delta$, the values \val f\in{\vektor{y}_1} and \val f\in{\vektor{y}_2}
        differ by less than $\varepsilon$.
\end{definition}

\noindent
Denoting $\delta  \defeq \omega^{-1}(\mapping{f}, \varepsilon)$, the general bound we obtained is
\begin{equation*}
\label{eqn:mainbound}
h \leq
\left(\frac{2e}{d} \left(\frac{\sqrt{\pi}^d\left(\frac{\diameter{\anset{K}}}{2}+\frac{\delta}{6}\right)^d}{\val \Gamma\in{\frac{d}{2}+1}}
  \left(\frac{2}{\varepsilon}\right)^C \!\!\!  +1\right)\right)^d\!\! ,
\end{equation*}
where
$$
C = \left[ 4 + \frac{3}{2}\log_2(d)\right]\frac{3\sqrt{d}\diameter{\anset{K}}}{\delta} \; .
$$

Artificial neural networks approximate functions described by data by
composing one-dimensional transition functions with the argument
$\scalarp{\vektor{w}}{\vektor{x}}$. Such functions are examples of so-called
ridge functions, which are constant on hyperplanes perpendicular to the vector
$\vektor{w}.$ This property makes it impossible to use naive and trivial
approaches to constructing approximating functions applicable when
approximating a function in a one-dimensional case or when approximating a
continuous function using, for example, RBF functions. One way to overcome the
problem of approximating a function on an n-dimensional compact using ridge
functions is to use the Stone-Weierstrass theorem, which states, among other
things, that the linear envelope of exponential functions with arguments
$\scalarp{\vektor{w}}{\vektor{x}}$ is dense in the space of functions
continuous on the compact. Then it suffices to approximate each exponential
function with argument $\scalarp{\vektor{w}}{\vektor{x}}$ by the sum of
transition functions of neurons again with argument
$\scalarp{\vektor{w}}{\vektor{x}}$, which is essentially an approximation task
in the one-dimensional case. Unfortunately, the standard proof of the SW
theorem, based on the approximation of the function using Tyler's expansion of
the function $\sqrt{1-t}$, does not provide a known possibility for estimating
the number of exponential functions with arguments
$\scalarp{\vektor{w}}{\vektor{x}}$ necessary for approximating the function
with a given accuracy. Of course, there are constructive proofs of various
forms of the SW theorem. However, even these are
not suitable for constructing and estimating the complexity of the
approximating function. For this, we need a different approach.

\bigskip

Our construction is inspired by an alternative proof of the Stone-Weierstrass
theorem, in particular, the proof by Brosowski and Deutsch
\cite{borowskydeutsch}, and it is elementarily constructive -- the proof
could be followed to obtain a network satisfying the bound.

We divide the graph of the target function \mapping{f} into horizontal slices
of height $\frac{\varepsilon}{2}$, approximate the indicator function of each
of these slices, and sum them to get an approximation of \mapping{f}. The
non-trivial step here is approximating the indicator functions of the slice
sets, as these can be complex and challenging to handle. Our solution is to
divide the compact \anset{K} into hypercubes, approximate the indicator
function of each of them, and take the product of those approximants that
correspond to cubes intersecting the set. The indicator function of a single
hypercube is then approximated by taking an exponential function for each of
its facets and carefully exponentiating their average

\section{Constructive Approximation of a Continuous Function}

We start with a formal definition of functions computable by a one-hidden-layer neural network with transfer function \mapping{\phi}.

\begin{definition}
	\label{def:GPhiEm}
	For \funct \psi \maps \Real{} \to{\Real{}} and  $m \in \Natural{}$ denote $\anset{G}_{\mapping{\psi}, m}$ as 
	\begin{equation*}
	\anset{G}_{\mapping{\psi}, m} \defeq \left\{\sum_{i=1}^m c_i\val \psi
          \in{\scalarp{\vektor{a}_i}{\vektor{x}}+b_i}\right\} \; ,
	\end{equation*}
where \ALL{i\leq m}\HOLDS{\vektor{a}_i \in \Real{d}, c_i,b_i \in \Real{}} . 
\end{definition}

Let us point out that $\anset{G}_{\mapping{exp}, m}$ denotes the set of
functions representable by a shallow neural network with the exponential
activation function, $\exp(t) = e^t$, that has $m$ units in the hidden layer.

The construction of a linear combination of exponential functions
approximating a given continuous function on a compact set in \Real{d} is
based on the approximation of the characteristic function of the
$d$-dimensional cube, which is elaborated in the following lemma.

%%%%%%%%%%%%%%%%%%%%%%%%%%%%%%%%%%%%%%%%%%%%%%%%%%%%%%%%%%%%%%%%%%%%%%%%%%%%%%%%%%%%%%%%%%%%%%%%%%%%%%%%%%%%%%%%%%%
%%%%%%% aproximace krychle %%%%%%%%%%%%%%%%%%%%%%%%%%%%%%%%%%%%%%%%%%%%%%%%%%%%%%%%%%%%%%%%%%%%%%%%%%%%%%%%%%%%%%%%
%%%%%%%%%%%%%%%%%%%%%%%%%%%%%%%%%%%%%%%%%%%%%%%%%%%%%%%%%%%%%%%%%%%%%%%%%%%%%%%%%%%%%%%%%%%%%%%%%%%%%%%%%%%%%%%%%%%

\begin{lemma}
  \label{lem:approxpolytope}
  Let $\epsilon\in\ooint{0}{1}$, $r > 0$, $\omega > 1$ and compact set
  $\anset{K} \subset \Real{d}$ is given. Further let $\vektor{x}_0 \in
  \Real{d}$ and assume that $\anset{I}(\vektor{x}_0, r) \cap \anset{K} \neq \emptyset$, where
  $\anset{I}\left(\vektor{x}_0, z\right) \defeq \vektor{x}_0 + \ccint{-z}{z}^d$, $z\in\Realp{}$. Then there exists \funct g\maps \Real{d}
  \to{\Real{}} such that
  \begin{enumerate}
    \item $\mapping{g}|_K: \anset{K} \rightarrow \ccint{0}{1}$,
    \item for $\vektor{y} \in \anset{I}(\vektor{x}_0, r) \cap
         \anset{K}$ is $\val g\in{\vektor{y}} > 1-\varepsilon$,
    \item for $\vektor{y} \in \anset{K} \minusset
         \anset{I}(\vektor{x}_0, \omega r)$ is  $\val g\in{\vektor{y}} < \varepsilon$,
    \item $\mapping{g} \in G_{\exp, h}$ for
    \begin{equation*}
          h = \left(\frac{2e}{d}\right)^d \left(\left(\frac{2}{\varepsilon}\right)^{\left[ 4 + \frac{3}{2}\log_2(d)\right]\frac{\diameter{\anset{K}}}{(\omega - 1)r} }+1\right)^d  .
    \end{equation*}
  \end{enumerate}
\end{lemma}
\beginproof
\proofpart{1),2),3}
If $\anset{K} \minusset I(\vektor{x}_0, \omega r)$ is empty, the claim can be satisfied by a constant function $Ce^{\scalarp{\vektor{0}}{\vektor{y}}}$, where $C\in\ooint{1-\epsilon}{1}$. Therefore in the next we assume that $\anset{K} \minusset I(\vektor{x}_0, \omega r)\neq \emptyset$.
\vspace{1ex}

\noindent
Denote $\vektor{x}_0 = \left(x_0^1, \dots, x_0^d \right)^\top$ and $s \defeq \frac{\log(4d)}{(\omega-1)r} > 0$.

\noindent
Let $\vektor{y} \in \anset{I}\left(\vektor{x}_0, r\right)$. Then for all \natur{i}{d} the following implication chain is true:
$$
\scalarp{\vektor{e}_i}{\vektor{y}}\in\ccint{\scalarp{\vektor{e}_i}{\vektor{x}_0}-r}{\scalarp{\vektor{e}_i}{\vektor{x}_0}+r}
\Rightarrow
$$
$$
\Rightarrow \scalarp{-\vektor{e}_i}{\vektor{y}} \leq -x_0^i+r \quad \mbox{and} \quad
  \scalarp{\vektor{e}_i}{\vektor{y}} \leq x_0^i+r \Rightarrow
$$
$$
 \Rightarrow \scalarp{-\vektor{e}_i}{\vektor{y}} +\left(x_0^i - \frac{r+\omega r}{2}\right)\leq
   \frac{(1-\omega)r}{2}  \quad \mbox{and}  
$$
$$
\scalarp{\vektor{e}_i}{\vektor{y}} -\left(x_0^i + \frac{r+\omega r}{2}\right)\leq \frac{(1-\omega)r}{2} 
\Rightarrow
$$
\bigskip
$$
 s\left[\scalarp{-\vektor{e}_i}{\vektor{y}} +\left(x_0^i - \frac{r+\omega r}{2}\right)\right]\leq
 -\frac{1}{2} log\left(4d\right)
$$
\centerline{and}  
$$
 s\left[\scalarp{\vektor{e}_i}{\vektor{y}} -\left(x_0^i + \frac{r+\omega r}{2}\right)\right]\leq  -\frac{1}{2} log\left(4d\right) 
\Rightarrow
$$
\bigskip
$$
 e^{\left(s\left[\scalarp{\vektor{e}_i}{\vektor{y}} +\left(x_0^i - \frac{r+\omega r}{2}\right)\right]\right)}\leq
 e^{-\frac{1}{2} log\left(4d\right)} = \frac{1}{2\sqrt{d}}
 $$
\centerline{and}  
$$
e^{\left(s\left[\scalarp{\vektor{e}_i}{\vektor{y}} -\left(x_0^i + \frac{r+\omega r}{2}\right)\right]\right)}\leq
 e^{-\frac{1}{2} log\left(4d\right)} = \frac{1}{2\sqrt{d}} \; .
$$
It follows that for all $\vektor{y} \in \anset{I}\left(\vektor{x}_0, r\right)$ is
$$
  \sum_{i=1}^{d} e^{s\left(\scalarp{-\vektor{e}_i}{\vektor{y}} + \left(x_0^i - \frac{r+\omega r}{2}\right)\right)} +
$$
$$
 + \sum_{i=1}^{d} e^{s\left(\scalarp{\vektor{e}_i}{\vektor{y}} - \left(x_0^i + \frac{r+\omega r}{2}\right)\right)} \leq
 \frac{2d}{2\sqrt{d}} = \sqrt{d}
$$
and finally

\begin{equationarray}{ll} 
  \sum_{i=1}^{d}  (4d)^{\frac{\left(\scalarp{-\vektor{e}_i}{\vektor{y}}
      + \left(x_0^i - \frac{r+\omega r}{2}\right)\right)}{(\omega-1)r}} \; + &    \nonumber \\
 + \sum_{i=1}^{d} (4d)^{\frac{\left(\scalarp{\vektor{e}_i}{\vektor{y}} - \left(x_0^i +
           \frac{r+\omega r}{2}\right)\right)}{(\omega-1)r}} & \leq   \sqrt{d}   \; . \label{eq:inexp}  \\ 
 \nonumber
\end{equationarray}

\noindent
Resembly, if $\vektor{y} \not\in \anset{I}\left(\vektor{x}_0, \omega r\right)$, then there exists \natur{j}{d} such that
the following implication chain is true:
$$
\scalarp{\vektor{e}_j}{\vektor{y}}\not\in
\ccint{\scalarp{\vektor{e}_j}{\vektor{x}_0}-\omega r}{\scalarp{\vektor{e}_j}{\vektor{x}_0}+\omega r}
\Rightarrow
$$
$$
\scalarp{-\vektor{e}_j}{\vektor{y}} > -x_0^j+\omega r
\quad
\mbox{or}
\quad
\scalarp{\vektor{e}_j}{\vektor{y}} > x_0^j+ \omega r \Rightarrow
$$
$$
 \Rightarrow \scalarp{-\vektor{e}_j}{\vektor{y}} +\left(x_0^j - \frac{r+\omega r}{2}\right) >
   \frac{(\omega - 1)r}{2}  \quad \mbox{or}  
$$
$$
\scalarp{\vektor{e}_j}{\vektor{y}} -\left(x_0^j + \frac{r+\omega r}{2}\right) > \frac{(\omega-1)r}{2} 
\Rightarrow
$$
$$
 \Rightarrow s\left[\scalarp{-\vektor{e}_j}{\vektor{y}} +\left(x_0^j - \frac{r+\omega r}{2}\right)\right] >
   \frac{1}{2} log\left(4d\right)    
$$
\centerline{or}
$$
s\left[\scalarp{\vektor{e}_j}{\vektor{y}} -\left(x_0^j + \frac{r+\omega r}{2}\right)\right] >  \frac{1}{2} log\left(4d\right) 
\Rightarrow
$$
$$
 \Rightarrow e^{\left(s\left[\scalarp{\vektor{e}_j}{\vektor{y}} +\left(x_0^j - \frac{r+\omega r}{2}\right)\right]\right)} >
 e^{\frac{1}{2} log\left(4d\right)} = 2\sqrt{d}   \quad \mbox{or}  
$$
$$
e^{\left(s\left[\scalarp{\vektor{e}_j}{\vektor{y}} -\left(x_0^j + \frac{r+\omega r}{2}\right)\right]\right)} >
 e^{\frac{1}{2} log\left(4d\right)} = 2\sqrt{d}  \; .
$$
It follows that for all $\vektor{y} \not\in \anset{I}\left(\vektor{x}_0, \omega r\right)$ is
\begin{equationarray*}{ll} 
  \sum_{i=1}^{d} e^{s\left(\scalarp{-\vektor{e}_j}{\vektor{y}} + \left(x_0^j -
        \frac{r+\omega r}{2}\right)\right)} \; + & \nonumber \\
 + \sum_{i=1}^{d} e^{s\left(\scalarp{\vektor{e}_j}{\vektor{y}} - \left(x_0^j +
       \frac{r+\omega r}{2}\right)\right)} &
 >  2\sqrt{d} \nonumber \\
\end{equationarray*}

\vspace{-2em}
\noindent
and finally
\begin{equationarray}{ll}
  \sum_{i=1}^{d} (4d)^{\frac{\left(\scalarp{-\vektor{e}_j}{\vektor{y}} +
        \left(x_0^j - \frac{r+\omega r}{2}\right)\right)}{(\omega-1)r}} \; + &   \nonumber \\
  + \sum_{i=1}^{d}  (4d)^{\frac{\left(\scalarp{\vektor{e}_j}{\vektor{y}} -
        \left(x_0^j + \frac{r+\omega r}{2}\right)\right)}{(\omega-1)r}} & > 2\sqrt{d} \; .  \label{eq:outexp} \\
  \nonumber
\end{equationarray}

\noindent
Define 
\begin{equationarray}{ll}
\val p\in{\vektor{x}} \defeq  &
\frac{1}{2d} \sum_{i=1}^{d}
  e^{s\left(\scalarp{-\vektor{e}_j}{\vektor{x}} + \left(x_0^j - \frac{r+\omega
          r}{2}\right)- \diameter{\anset{K}} \right)} \; + \nonumber \\
  & \frac{1}{2d} \sum_{i=1}^{d}e^{s\left(\scalarp{\vektor{e}_j}{\vektor{x}} - \left(x_0^j +
        \frac{r+\omega r}{2}\right)- \diameter{\anset{K}} \right)}   , \;
  \; \label{DefinitionOfPx} \\
\nonumber
\end{equationarray}
where \diameter{\anset{K}} is the diameter of \anset{K}.  The mapping
\mapping{p} is clearly non-negative.

\noindent
Further let $\vektor{y} \in \anset{K}$ and $\vektor{y}^\prime \in I(\vektor{x}_0, r) \cap \anset{K}$ ($\neq \emptyset$).
Obviously for any \natur{j}{d} is
$$
\left(x_0^j - \frac{r+\omega r}{2}\right) \leq \scalarp{\vektor{e}_j}{\vektor{y}^\prime} \leq \left(x_0^j + \frac{r+\omega r}{2}\right) \; .
$$
So we have for any $\vektor{y} \in \anset{K}$ and \natur{j}{d} 
\begin{equationarray}{c}
  \scalarp{-\vektor{e}_j}{\vektor{y}} + \left(x_0^j - \frac{r+\omega
      r}{2}\right) \leq \nonumber \\ \scalarp{-\vektor{e}_j}{\vektor{y}} +
  \scalarp{\vektor{e}_j}{\vektor{y}^\prime}  
  < \norm{\vektor{y}  - \vektor{y}^\prime}{}\leq \diameter{\anset{K}} \label{DefinitionOfPxA} \\
 \nonumber
\end{equationarray}
and 
\begin{equationarray}{c}
  \scalarp{\vektor{e}_j}{\vektor{y}} - \left(x_0^j + \frac{r+\omega
      r}{2}\right) \leq \nonumber \\ \scalarp{\vektor{e}_j}{\vektor{y}} -
  \scalarp{\vektor{e}_j}{\vektor{y}^\prime} < \norm{\vektor{y} -
    \vektor{y}^\prime}{}\leq \diameter{\anset{K}} \; . \label{DefinitionOfPxB} \\
  \nonumber
\end{equationarray}
Thus, each summand in the expression \ref{DefinitionOfPx} is less than 1. Hence $\mapping{p} \in \ooint{0}{1}$ on \anset{K}.

\vspace{1em}

\noindent
Using the definition of $s$ we can rewrite the \val p\in{\vektor{x}} as
\begin{equationarray}{ll}
  \val p\in{\vektor{x}} = \! \! \! & 
  \frac{1}{2d} \sum_{i=1}^{d} (4d)^\frac{\scalarp{-\vektor{e}_j}{\vektor{x}} + 
      \left(x_0^j - \frac{r+\omega r}{2}\right)- \diameter{\anset{K}}}{(\omega-1 )r} +  \nonumber \\
  &  \frac{1}{2d} \sum_{i=1}^{d}
  (4d)^\frac{\scalarp{\vektor{e}_j}{\vektor{x}} - \left(x_0^j + \frac{r+\omega
        r}{2}\right)- \diameter{\anset{K}}}{(\omega -1 )r} \; .  \nonumber \\
\nonumber
\end{equationarray}

\begin{figure}[h]
\centerline{
\resizebox{0.47\textwidth}{!}{
  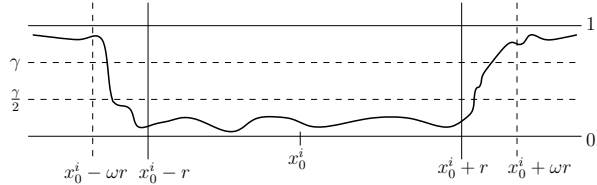}
}
    \caption[Local approximation.]{Local approximation on the cube.}
    \label{SumOfReLUGi}
\end{figure}

\noindent
Let $\vektor{y} \in \anset{I}\left(\vektor{x}_0, r\right)$, then the estimation \ref{eq:inexp} follows
$$
\val p\in{\vektor{y}} \leqfrom{\ref{eq:inexp}} \frac{\sqrt{d}}{2d} \cdot (4d)^{-\frac{\diameter{\anset{K}}}{(\omega -1 )r}} = \frac{\gamma}{2} \; ,
$$

\noindent
and resembly, if $\vektor{y} \not\in \anset{I}\left(\vektor{x}_0, \omega r\right)$, then the estimation \ref{eq:outexp} follows
$$
\val p\in{\vektor{y}} \geqfrom{\ref{eq:outexp}} \frac{2\sqrt{d}}{2d} \cdot (4d)^{-\frac{\diameter{\anset{K}}}{(\omega -1 )r}}  = \gamma \; ,
$$
where we denote 
\begin{equation*}
\label{DefinitionOfGamma}
  \gamma \defeq \frac{1}{\sqrt{d}} (4d)^{-\frac{\diameter{\anset{K}}}{(\omega -1)r}} \; .
\end{equation*}
Figure \ref{SumOfReLUGi} sketches those properties of the function \mapping{p}.

\noindent
By the claim assumption and the starting note of the proof, we have
$I(\vektor{x}_0, r) \cap \anset{K} \neq \emptyset$ and
$\anset{K} \minusset I(\vektor{x}_0, \omega r) \neq \emptyset$. Hence there
exists $\vektor{z}\in I(\vektor{x}_0, r) \cap \anset{K}$ and
$\vektor{z}^\prime \in \anset{K} \minusset I(\vektor{x}_0, \omega r)$ such
that $\norm{\vektor{z} - \vektor{z}^\prime}{} > (\omega - 1)r$. It follows
$\diameter{\anset{K}} > (\omega - 1)r$ and $\gamma$ as a function of $d$ is
monotonically decreasing and for any $d\geq 1$ is $\gamma < 1$.  Put
$$
\alpha \defeq \frac{\gamma}{2} \quad \mbox{and} \quad \beta \defeq \gamma \; . 
$$
Then
$$
\gamma < 1 \Leftrightarrow \frac{2}{\gamma} - \frac{1}{\gamma} > 1
\Leftrightarrow \frac{1}{\alpha} - \frac{1}{\beta} > 1 \; . 
$$

\noindent
Finally, define the function $\mapping{g}$ as
\begin{equationarray}{lll}
 \val g\in{\vektor{x}} & \!\defeq \!\!\! & \left(1 -\left[\val p\in{\vektor{x}} \right]^n
 \right)^{k^n} = \nonumber  \\
& \! = \! & \left(\!1 \!-\! \left[
 \frac{1}{2d} \!\sum_{i=1}^{d} \!\left( (4d)^{\cal S} + (4d)^{\cal T}
 \right)\right]^n \right)^{k^n} \!\!\!\! , \;\;\;\;\label{DefinitionGonCube}\\
\nonumber
\end{equationarray}

where
$$
{\cal S} = \frac{\scalarp{-\vektor{e}_j}{\vektor{x}} + \left(x_0^j - \frac{r+\omega
      r}{2}\right)- \diameter{\anset{K}}}{(\omega -1 )r} \; , 
$$
$$
{\cal T} = \frac{\scalarp{\vektor{e}_j}{\vektor{x}} - \left(x_0^j + \frac{r+\omega
         r}{2}\right)- \diameter{\anset{K}}}{(\omega -1 )r}   
$$
and $n = \left\lceil \frac{-\log(\varepsilon)}{\log(2)} \right\rceil$, $k = \left\lfloor \sqrt{2d} (4d)^{{}^{\frac{\diameter{\anset{K}}}{(\omega - 1)r}}} \right\rfloor+1$. As $n$ and $k$ are both positive integers, it follows that $\mapping{g} \in G_{\exp,m}$ for some natural $m$. We have shown (see \ref{DefinitionOfPx},
\ref{DefinitionOfPxA} and \ref{DefinitionOfPxB}) that $\mapping{g}|_K$ ranges in \ccint{0}{1}. To prove statements (2) and (3) of the lemma proved, recall that $\alpha=\frac{\gamma}{2}$, $\beta = \gamma$, $n$ and $k$ defined above meet assumptions of the lemma \ref{ApplicationOof_BernoulliInequality}  and hence for $\vektor{y} \in I(\vektor{x}_0, r) \cap \anset{K}$ is $\val g\in{\vektor{y}} > 1-\varepsilon$ and for $\vektor{y} \in \anset{K} \minusset I(\vektor{x}_0, \omega r)$ is
$\val g\in{\vektor{y}} < \varepsilon$.

\vspace{1em}

\proofpart{4}
As for the number of hidden units, we know that $\mapping{p} \in G_{\exp,m}$. By multinomial theorem the number of terms in
$\mapping{g} = \left(1-\mapping{p}^n\right)^{k^n}$ is at most the number of terms in $\left(1-\mapping{p}\right)^{nk^n}$. Therefore, by Lemma
\ref{lem:cubeexp} $\mapping{g} \in G_{\exp, h}$ for
\begin{equation}
\label{Auxiliary_HEstimation}
  h < \left( 2e \left( \frac{nk^n}{d} + 1 \right) \right)^d \; .
\end{equation}

\noindent
Furthermore,
\begin{equation}
  \label{Auxiliary_NEstimation}  
	n \leq -\frac{\log( \varepsilon)}{\log(2)} =  \log_2 \left(\frac{2}{\varepsilon}\right)
\end{equation}
and (recall $\diameter{\anset{K}} > (\omega - 1)r$)
\begin{equationarray*}{ll}
  k & \leq \sqrt{2d} (4d)^{{}^{\frac{\diameter{\anset{K}}}{(\omega - 1)r}}} +1 <
  \sqrt{2d} (5d)^{{}^{\frac{\diameter{\anset{K}}}{(\omega - 1)r}}}  = \\
& = 2^{{}^{  \frac{1 +\log_2(d)}{2} + \left(\log_2(5)
       +\log_2(d)\right)\frac{\diameter{\anset{K}}}{(\omega - 1)r} }} < \\
& < 2^{{}^{ \left[ \frac{1 +\log_2(d)}{2} +
      \log_2(5)+\log_2(d)\right]\frac{\diameter{\anset{K}}}{(\omega - 1)r} }} < \\
& <  2^{{}^{ \left[ 3 + \frac{3}{2}\log_2(d)\right]\frac{\diameter{\anset{K}}}{(\omega - 1)r} }} \defeq \lambda \; .
\end{equationarray*}
Using equivalency $k < \lambda \Leftrightarrow k < 2^{\log_2(\lambda)}$ we have
\begin{equation}
  \label{Auxiliary_KEstimation}
  k^n < 2^{n\log_2(\lambda)} \leq   2^{\log_2 \left(\frac{2}{\varepsilon}\right)\log_2(\lambda)}
  =  \left(\frac{2}{\varepsilon}\right)^{\log_2(\lambda)} 
\end{equation}
and therefore by \ref{Auxiliary_HEstimation}, \ref{Auxiliary_NEstimation} and \ref{Auxiliary_KEstimation} we derive
\begin{equationarray*}{ll}
  h & < \left(\frac{2e}{d} \right)^d \left(
    \log_2\left(\frac{2}{\varepsilon}\right)\left(\frac{2}{\varepsilon}\right)^{\log_2(\lambda)}
    +1\right)^d  < \\
& < \left(\frac{2e}{d}\right)^d
\left(\left(\frac{2}{\varepsilon}\right)^{1+\log_2(\lambda)}+1 \right)^d  = \\
& =  \left(\frac{2e}{d}\right)^d
\left(\left(\frac{2}{\varepsilon}\right)^{1+\left[ 3 +
      \frac{3}{2}\log_2(d)\right]\frac{\diameter{\anset{K}}}{(\omega - 1)r}
  }+1\right)^d  \leq \\
& \leq\left(\frac{2e}{d}\right)^d
\left(\left(\frac{2}{\varepsilon}\right)^{\left[ 4 +
      \frac{3}{2}\log_2(d)\right]\frac{\diameter{\anset{K}}}{(\omega - 1)r}
  }+1\right)^d  \; . \\
\end{equationarray*}
\vspace{-2em}
\endproof

%%%%%%%%%%%%%%%%%%%%%%%%%%%%%%%%%%%%%%%%%%%%%%%%%%%%%%%%%%%%%%%%%%%%%%%%%%%%%%%%%%%%%%%%%%%%%%%%%%%%%%%%%%%%%%%%%%%
%%%%%%% odhad normy f - g %%%%%%%%%%%%%%%%%%%%%%%%%%%%%%%%%%%%%%%%%%%%%%%%%%%%%%%%%%%%%%%%%%%%%%%%%%%%%%%%%%%%%%%%%
%%%%%%%%%%%%%%%%%%%%%%%%%%%%%%%%%%%%%%%%%%%%%%%%%%%%%%%%%%%%%%%%%%%%%%%%%%%%%%%%%%%%%%%%%%%%%%%%%%%%%%%%%%%%%%%%%%%

The importance of the lemma just proved lies in the fact that it allows local
approximation of a given function by a linear combination of exponential
functions with argument $\scalarp{\vektor{w}}{\vektor{x}}$. As can be seen
from the claim of the proven lemma, the number of exponentials necessary for
the approximation depends on the value of the approximation accuracy
$\epsilon$ and the value of $\omega$, which determines the size of the
transition region. For the purpose of approximating a given continuous
function on the entire compact \anset{K}, we divide the compact \anset{K} into
a network of adjacent cubes, and the value of the function on each cube we
approximate by the sum of exponentials according to the previous lemma. Since
the accuracy of the approximation $\epsilon$ is a value we have chosen, it
remains to specify the value of the parameter $\omega$, which, as we will see
later, will depend on the inverse modulus of continuity of the approximated
function. This construction and the estimate of the number of necessary
exponentials are the subject of the following statement.

\begin{theorem}
\label{thm:fapprox}
Let $\anset{K} \subset \Real{d}$ be compact, \funct f\maps
\anset{K}\to{\Real{}} be continuous and $\epsilon \in
\ooint{0}{\frac{1}{2}}$. Denote by
$\delta \defeq \omega^{-1}\left(\mapping{f}, \frac{\epsilon}{2}\right)$ the
inverse of the modulus of continuity of \mapping{f} at $\frac{\epsilon}{2}$.
Then there exists $\mapping{g} \in G_{\exp, h}$ such that
$$ \norm{\mapping{f} - \mapping{g}}{\infty} < 2\epsilon
$$
and    
\begin{equation*}
  h \leq
\left(\frac{2e}{d} \left(\frac{\sqrt{\pi}^d\left(\frac{\diameter{\anset{K}}}{2}+\frac{\delta}{6}\right)^d}{\val \Gamma\in{\frac{d}{2}+1}}
  \left(\frac{2}{\varepsilon}\right)^C\!\!\!\!+1\right)\right)^d  \!\! ,
\end{equation*}
where
$$
C = \left[ 4 + \frac{3}{2}\log_2(d)\right]\frac{3\sqrt{d}\diameter{\anset{K}}}{\delta} \; .
$$
\end{theorem}

\beginproof
\proofpart{prove the existence of \mapping{g}}
Let $r \defeq \frac{\delta}{3\sqrt{d}}$ and take $\anset{X} \defeq r \Integers{d}$. Denoting again
$I(\vektor{x}, r) = \vektor{x} + \ccint{-r}{r}^d$, so for all $\vektor{x} \in \anset{X}$ the inradius of the hypercube $I\left(\vektor{x}, r\right)$
is $r$ and its circumradius is $\frac{\delta}{3}$. Define further
$\anset{X}_{\anset{K}} \defeq \defset{\vektor{x} \in \anset{X}}{I\left(\vektor{x}, r\right) \cap \anset{K} \neq \emptyset}$ and for all
$\vektor{x} \in \anset{X}_{\anset{K}}$ put
$$
v_{\vektor{x}}^{max} \defeq  \max_{\vektor{y} \in I\left(\vektor{x},   r\right)}  \left\{ \val f\in{\vektor{y}} \right\},
$$
$$
v_{\vektor{x}}^{min} \defeq   \min_{\vektor{y} \in I\left(\vektor{x}, r\right)}  \left\{ \val f\in{\vektor{y}} \right\}, 
$$
$$
v_{\vektor{x}}^{mid} \defeq \frac{1}{2}\left(   v_{\vektor{x}}^{min} +  v_{\vektor{x}}^{max}\right)\; .  
$$
The continuity of \mapping{f} and compactness of
$I\left(\vektor{x}, r\right)$ follow that there exist $\vektor{y}_{\vektor{x}}^{max},
\vektor{y}_{\vektor{x}}^{min}, \vektor{y}_{\vektor{x}}^{mid} \in I\left(\vektor{x}, r\right)$  such that
$v_{\vektor{x}}^{min}=\val f\in{\vektor{y}_{\vektor{x}}^{min}}$,
$v_{\vektor{x}}^{max}=\val f\in{\vektor{y}_{\vektor{x}}^{max}}$ and
$v_{\vektor{x}}^{mid}=\val f\in{\vektor{y}_{\vektor{x}}^{mid}}$.

\noindent
Obviously, the diameter of the set $I\left(\vektor{x}, 2r\right)$ is
$\frac{2}{3}\delta < \delta$ and therefore, by the definition of inverse modulus
of continuity,  $v_{\vektor{x}}^{max} -
v_{\vektor{x}}^{min} \leq \frac{\epsilon}{2}$, so we have 

\begin{equation}
\label{yInR}  
\vektor{y} \in I\left(\vektor{x},r\right) \Rightarrow
\absvalue{\val f\in{\vektor{y}} - v_{\vektor{x}}^{mid}} \leq \frac{\epsilon}{4} \; .
\end{equation}

\noindent
Now assume that $\vektor{y} \in I\left(\vektor{x}, 2r\right) \minusset I\left(\vektor{x}, r\right)$:

\noindent
It is straightforward that the circumradius of
$I\left(\vektor{x}, 2 r\right)$ is $\frac{2\delta}{3}$. So
\begin{equation*}
  \norm{\vektor{y} - \vektor{y}_{\vektor{x}}^{mid}}{} \leq \norm{\vektor{y} - \vektor{x}}{} + \norm{\vektor{x} -
    \vektor{y}_{\vektor{x}}^{mid}}{} \leq \frac{2}{3}\delta + \frac{1}{3}\delta = \delta \; , 
\end{equation*}
hence by assumption of the theorem 
\begin{equation}
\label{yIn2RAndNotInR}  
\vektor{y} \in I\left(\vektor{x}, 2r\right) \minusset I\left(\vektor{x}, r\right)
\Rightarrow
\absvalue{\val f\in{\vektor{y}} - v_{\vektor{x}}^{mid}} \leq \frac{\epsilon}{2} \; .  
\end{equation}

\noindent
Let's  analyze the case when $\vektor{y} \in I\left(\vektor{x}, 2r\right)
\minusset I\left(\vektor{x}, r\right)$ and $\vektor{y} \in
I\left(\vektor{x}^\prime, r\right)$, where $\vektor{x} \neq
\vektor{x}^\prime$: 

\noindent
In this case, by \ref{yInR} is $\absvalue{\val f\in{\vektor{y}} - v_{\vektor{x}^\prime}^{mid}} \leq \frac{\epsilon}{4}$ and by \ref{yIn2RAndNotInR} is
$\absvalue{\val f\in{\vektor{y}} - v_{\vektor{x}}^{mid} }\leq \frac{\epsilon}{2}$. Both together give
\begin{equationarray}{c}
\vektor{y} \in I\left(\vektor{x}, 2r\right) \minusset I\left(\vektor{x},
  r\right) \; \mbox{and} \; \vektor{y} \in I\left(\vektor{x}^\prime, r\right) \nonumber \\[1mm]
\mbox{and} \; \vektor{x} \neq \vektor{x}^\prime \quad \Rightarrow \quad
\absvalue{v_{\vektor{x}}^{mid} - v_{\vektor{x}^\prime}^{mid}} \leq
  \frac{3}{4}\epsilon \; . \label{VxVxPrimeDistance} \\ 
 \nonumber
\end{equationarray}
So for all $\vektor{x},\vektor{x}^\prime\in\anset{X}$ holds
\begin{equation*}
\label{EmptyIntersectionOfIrI2r}  
\absvalue{v_{\vektor{x}}^{mid} - v_{\vektor{x}^\prime}^{mid}} \geq \epsilon \; \Rightarrow \;
 I\left(\vektor{x}, 2r\right) \cap I\left(\vektor{x}^\prime, r\right) = \emptyset \;  .  
\end{equation*}

\noindent
Denote
$$
k \defeq \absvalue{\anset{X}_{\anset{K}}}
$$
and
$$
n \defeq \left\lfloor \frac{2}{\epsilon}\left(\max_{\vektor{y} \in \anset{K}}  \left\{ \val f\in{\vektor{y}} \right\} -
    \min_{\vektor{y} \in \anset{K}}  \left\{ \val f\in{\vektor{y}} \right\} \right) \right\rfloor -1 \; .
$$

\noindent
Then for all $\vektor{x} \in \anset{X}_{\anset{K}}$ define $\mapping{g}_{\vektor{x}}$ as in
lemma \ref{lem:approxpolytope} such that $\mapping{g}_{\vektor{x}}|_K : \anset{K} \rightarrow \ccint{0}{1}$ and
\begin{equationarray*}{lcl}
\mapping{g}_{\vektor{x}}\left(\vektor{y}\right) > 1-\frac{\epsilon}{kn} & \mbox{if} & 
\vektor{y} \in I\left(\vektor{x}, r\right) \cap \anset{K}  \; \\
 \mapping{g}_{\vektor{x}}\left(\vektor{y}\right) < \frac{\epsilon}{kn}
 & \mbox{if} &  \vektor{y} \in \anset{K} \minusset I(\vektor{x}, 2r) \; . \\
\end{equationarray*}

\noindent
Further, for all  \natur{i}{n} define
$$
\anset{X}_i \defeq \defset{\vektor{x} \in \anset{X}_{\anset{K}}}{
  v_{\vektor{x}}^{mid} \geq i\epsilon}\; ,
$$
(so the sets $\anset{X}_i$ are ordered by inclusion, $\anset{X}_{i+1}
\subseteq \anset{X}_i$, \\ \natur{i}{n-1}) and define \funct p_i \maps
\anset{\Real{d}}\to{\Real{}} as
\begin{equation*}
 \mapping{p}_i \defeq 1 - \prod_{\vektor{x} \in \anset{X}_i} \left(1 - \mapping{g}_{\vektor{x}}\right) \; .
\end{equation*}

\noindent      
Obviously $\mapping{p}_i|_{\anset{K}} : \anset{K} \rightarrow \ccint{0}{1}$. Taking advantage of the fact that
$\bigcup_{\vektor{x} \in \anset{X}_i} I\left(\vektor{x}, r\right) \subset \bigcup_{\vektor{x} \in \anset{X}_i} I\left(\vektor{x}, 2r\right)$
we can distinguish three disjunctive cases (see Lemma \ref{Disjunctive_Sets_Lemma}):

\bigskip

\begin{enumerate}
\item $\vektor{y} \in \anset{K} \cap \left[ \bigcup_{\vektor{x} \in \anset{X}_i} I\left(\vektor{x}, r\right)\right]\; :$

\noindent
In this case there exists one $\vektor{x}_* \in \anset{X}_i$ such that $\vektor{y} \in I\left(\vektor{x}_*, r\right)$, so
$\mapping{g}_{\vektor{x}_*}\left(\vektor{y}\right) > 1-\frac{\epsilon}{kn}$. In addition, for all $\vektor{x} \in \anset{X}_i$ is
$1-\mapping{g}_{\vektor{x}}\left(\vektor{y}\right) < 1$. Hence
\begin{equationarray}{ll}
  \mapping{p}_i\left(\vektor{y}\right)\! = & \! 1\! - \!\left(1\! -\! 
    \mapping{g}_{\vektor{x}_*}\left(\vektor{y}\right)\right)\left(
    \prod_{{\vektor{x} \in
      \anset{X}_i \atop \vektor{x}\neq\vektor{x}_*}}\!\!\!\! \left(1 \!- \!\mapping{g}_{\vektor{x}}\left(\vektor{y}\right)\right)\!\!\right)>
\nonumber \\
\label{IsIn1rSquare} 
& > \mapping{g}_{\vektor{x}_*}\left(\vektor{y}\right) > 1 -  \frac{\epsilon}{kn}  \geq 1-\frac{\epsilon}{n} \; .
\end{equationarray}
      
\noindent
\item $\vektor{y} \in \anset{K} \minusset \bigcup_{\vektor{x} \in \anset{X}_i} I\left(\vektor{x}, 2r\right)\; :$

\noindent
In this case $1 - \mapping{g}_{\vektor{x}}\left(\vektor{y}\right) > 1 - \frac{\epsilon}{kn}$ for all $\vektor{x} \in \anset{X}_i$, so by Bernoulli's
inequality (see \ref{BernoulliMinus})
\begin{equationarray}{ll}
\val p_i\in{\vektor{y}} = &\!\!\!\! 1 \!-\! \prod_{\vektor{x} \in \anset{X}_i}\!\!\!
\left(1\! - \!\mapping{g}_{\vektor{x}}\left(\vektor{y}\right)\right) \!\! <
1\! -\! \left(\! 1\! - \!\frac{\epsilon}{kn}\right)^{\absvalue{\anset{X}_i}}\leqfrom{\ref{BernoulliMinus}} 
\nonumber \\
& \leqfrom{\ref{BernoulliMinus}}  
1 - 1 +   \frac{\epsilon}{n}\frac{\absvalue{\anset{X}_i}}{k} \leq
\frac{\epsilon}{n} \; . \label{NotIn2rSquare}\\ \nonumber
\end{equationarray}

\item
  $\vektor{y} \in \anset{K} \cap \left[ \bigcup_{\vektor{x} \in \anset{X}_i} I\left(\vektor{x}, 2r\right) \minusset \bigcup_{\vektor{x} \in
      \anset{X}_i} I\left(\vektor{x}, r\right)\right] \; :$

\noindent
In this case there exists at least one $\vektor{x} \in \anset{X}_i$ such that
$\vektor{y} \in \anset{K} \cap \left[I\left(\vektor{x}, 2r\right) \minusset I\left(\vektor{x}, r\right)\right]$. Then by \ref{yIn2RAndNotInR} is
$\absvalue{\val f\in{\vektor{y}} - v_{\vektor{x}}^{mid}} \leq
\frac{\epsilon}{2}$ .
\end{enumerate}  

\bigskip

\noindent
Finally define the function \funct g \maps \anset{\Real{d}}\to{\Real{}} as
\begin{equation*}
  \label{Definition_of_G_Function}
 \val g \in{\vektor{y}} \defeq \epsilon\sum_{j=1}^{n} \val p_j\in{\vektor{y}}  \; .
\end{equation*}

\noindent
Let $\vektor{y} \in \anset{K}$ be arbitrary and let $\vektor{x}_y \in \anset{X}_{\anset{K}}$ such that $\vektor{y} \in I\left(\vektor{x}_y, r\right)$, and
assume that $v_{\vektor{x}_y}^{mid} \in \coint{m\epsilon}{(m+1)\epsilon}$. It means that $\vektor{x}_y \in \anset{X}_i$ for \natur{i}{m} and
$\vektor{x}_y \notin \anset{X}_j$ for $j\geq m+1$. So by \ref{IsIn1rSquare} for all \natur{i}{m} is
$\mapping{p}_i\left(\vektor{y}\right) \geq 1-\frac{\epsilon}{n}$ and recall that for all \natur{i}{n} is
$\mapping{p}_i\left(\vektor{y}\right) \in \ccint{0}{1}$. These facts allow us to establish the lower estimate of the value of \val g \in{\vektor{y}}
as
\begin{equationarray}{ll}
  \val g \in{\vektor{y}} & = \epsilon\sum_{i=1}^m \val p_i\in{\vektor{y}}  + \epsilon\sum_{j = m+1}^n \val p_j\in{\vektor{y}}
  \geq  \nonumber \\
  & \geq \epsilon\sum_{i=1}^m \val p_i\in{\vektor{y}}  \geq  m\epsilon \left(1-\frac{\epsilon}{n}\right)   
  \geq  m\epsilon - \epsilon^2 \; . \nonumber \\
\nonumber
\end{equationarray}

\noindent
Resembly by \ref{NotIn2rSquare}, all $j>m+1$ imply $\mapping{p}_j\left(\vektor{y}\right) \leq \frac{\epsilon}{n}$. So we cas establish upper estimate
of the value of \val g \in{\vektor{y}} as
\begin{equationarray}{ll}
  \val g\in{\vektor{y}} & = \epsilon\sum_{i=1}^{m+1} \val p_i\in{\vektor{y}}  + \epsilon\sum_{j = m+2}^n \val p_j\in{\vektor{y}}
\leq \nonumber \\
&  \leq \epsilon\left( m+1 \right)  + \epsilon \left(n-m-2 \right)\frac{\epsilon}{n} \leq \left(m+1\right) \epsilon + \epsilon^2 \; .
 \nonumber \\ \nonumber 
\end{equationarray}

\noindent
That is, $\val g\in{\vektor{y}} \in \ccint{m\epsilon - \epsilon^2}{(m+1)\epsilon + \epsilon^2}$. Because
$v_{\vektor{x}_y}^{mid} \!\!\in \!\!\coint{m\epsilon}{(m+1)\epsilon}$, we get
$\absvalue{v_{\vektor{x}_y}^{mid} \!-\! \val g\in{\vektor{y}}} \!\! < \!\!\epsilon
\! +\! \epsilon^2$. We also
know that $\absvalue{v_{\vektor{x}_y}^{mid} - \mapping{f}(\vektor{y})} \leq
\frac{\epsilon}{2}$, hence ($\epsilon < \frac{1}{2}$ according to assumption)
\begin{equation*}
	\absvalue{\mapping{f}(\vektor{y}) - \val g\in{\vektor{y}}} <
        \epsilon+\epsilon^2+\frac{\epsilon}{2} = \epsilon\left(1+\epsilon+\frac{1}{2} \right)< 2\epsilon  \; . 
\end{equation*}
The vector $\vektor{y}\in\anset{K}$ was chosen arbitrarily, so finally we have
$\norm{\mapping{f} - \mapping{g}}{\infty} < 2\epsilon$.

\bigskip

\noindent
\proofpart{estimation of a sufficient number of exponentials}

\noindent
By the proof of the lemma
\ref{lem:approxpolytope} (see \ref{DefinitionGonCube}), $g_{\vektor{x}}$ is for all $\vektor{x} \in \anset{X}_{\anset{K}}$ of the form
\begin{equation*}
g_{\vektor{x}}\left(\vektor{y} \right)\! = \!\left(\! 1\! - \!\left[
 \frac{1}{2d} \sum_{i=1}^{d} \!\left( e^{t_j\ln\left(4d\right)} + e^{u_j\ln\left(4d\right)} \right)\!\right]^\nu \right)^{\kappa^\nu} \!\!\!\!\! , 
\end{equation*}
where (see \ref{Auxiliary_NEstimation}, \ref{Auxiliary_KEstimation} and recall
that $\omega = 2$ in this case) 
\begin{equation*}
 \label{NuEstimation} 
\nu \leq  \frac{-\log(\varepsilon)}{\log(2)} = \log_2 \left(\frac{2}{\varepsilon}\right)  \; ,
\end{equation*}
\begin{equation*}
 \label{KappaEstimation} 
 \kappa \!= \!\left\lfloor \!\sqrt{2d} (4d)^{{}^{\frac{\diameter{\anset{K}}}{r}}} \!\right\rfloor+1 \; , 
   \kappa^\nu \!\! < \!\left(\!\frac{2}{\varepsilon}\!\right)^{{ \!\left[ 3 + \frac{3}{2}\log_2(d)\right]\frac{\diameter{\anset{K}}}{r} }} 
\end{equation*}
and
$$
t_j = \frac{\scalarp{-\vektor{e}_j}{\vektor{y}} + \left(\vektor{x}_j -
    \frac{3r}{2}\right)- \diameter{\anset{K}}}{r} \; ,
$$
$$
u_j = \frac{\scalarp{\vektor{e}_j}{\vektor{y}} - \left(\vektor{x}_j + \frac{3r}{2}\right)- \diameter{\anset{K}}}{r}   \; .
$$
Recall that,
\begin{equation*}
  \frac{1}{\epsilon} \val g\in{\vektor{y}} = \sum_{j=1}^n \left( 1 -
    \prod_{\vektor{x} \in \anset{X}_j} \left(1- g_{\vektor{x}} \right) \right) \; . 
\end{equation*}
Hence, all exponential terms in $\frac{1}{\epsilon}\widetilde{g}$ are also contained in the expression 
\begin{equation*}
 \prod_{\vektor{x} \in \anset{X}} \left[1\!- \!\left(\!1 \!- \!
        \left[\frac{1}{2d}\sum_{i=1}^{d}  \left(e^{t_j\ln\left(4d\right)} + e^{u_j\ln\left(4d\right)}\! \right)  \!  \right]^{\nu}\right)^{\!\!\!\kappa^{\nu}}\right]  
\end{equation*}
which contains at most as many exponential terms as
\begin{equation}
	\label{eq:ch2expression}
	\left(1 - \sum_{i=1}^{d}\left( e^{t_j\ln\left(4d\right)} + e^{u_j\ln\left(4d\right)} \right)
        \right)^{\nu\kappa^{\nu}\setsize{\anset{X}_{\anset{K}}}} .
\end{equation}

\noindent
By Lemma \ref{lem:cubeexp}, the number of terms in (\ref{eq:ch2expression}) is at most
\begin{equation*}
   h < \left( 2e \left( \frac{\nu\kappa^{\nu}\setsize{\anset{X}_{\anset{K}}}}{d} + 1 \right) \right)^d   .
\end{equation*}
Using the claim of the lemma \ref{NumOfIntVectorsInBall}, 
\begin{equation*}
\setsize{\anset{X}_{\anset{K}}} \leq \frac{\sqrt{\pi}^d\left(\frac{\diameter{\anset{K}}}{2}+r\frac{\sqrt{d}}{2}\right)^d}{\val \Gamma\in{\frac{d}{2}+1}} \; ,
\end{equation*}
we can directly derive by substitution
\begin{equation}
\label{FinalMainResult} 
  h \leq
\left(\frac{2e}{d} \left(\frac{\sqrt{\pi}^d\left(\frac{\diameter{\anset{K}}}{2}+\frac{\delta}{6}\right)^d}{\val \Gamma\in{\frac{d}{2}+1}}
  \left(\frac{2}{\varepsilon}\right)^C\!\!\!\!+1\right) \right)^d  \!\! ,
\end{equation}
where
$$
C = \left[ 4 + \frac{3}{2}\log_2(d)\right]\frac{3\sqrt{d}\diameter{\anset{K}}}{\delta} \; .
$$

\endproof

\section{Conclusion}

The previous statement \ref{thm:fapprox} is one of the few constructive proofs
of the approximation properties of shallow neural networks that provides an
upper bound on the sufficient size of a shallow neural network. The basic
motivation for obtaining such a result is the need to design a neural network
of a suitable size for a given task, which would be large enough to ensure the
required approximation accuracy on the one hand, and on the other hand would
at least partially eliminate the effect of overfitting of the neural network
due to an excessive number of parameters.

However, the practical applicability of the estimate obtained by equation
\ref{FinalMainResult} is low for two reasons. The first reason is that it is a
general estimate valid for every continuous function on a compact set, and it
is difficult to expect that this estimate will be accurate for a specific
function describing the task at hand, which is a natural price to pay for the
generality of the obtained estimate. The second reason is the description of
the complexity of the approximated function using the continuity modulus,
which is a quantity dependent on the local behavior of the function, where two
almost identical functions differing only locally can have very different
values of the continuity modulus.

The continuity module is used only in statement \ref{thm:fapprox} to derive an
estimate of the size of a shallow neural network; we can adapt this statement
to use a different description of the curve of the approximated function. The
authors consider the global variation of the function, which depends on the
behavior of the approximated function throughout its domain, to be a suitable
candidate for this description.

The estimate in statement \ref{thm:fapprox} is for a shallow neural network
with one hidden layer, in which the neurons' transition functions are
exponential. The arguments of these exponential functions are scalar products
of the input value vector and weight vectors, so they are effectively
one-dimensional functions that are constant on hyperplanes perpendicular to
the weight vector. Thus, on the axis given by the linear envelope of the
weight vector, they are one-dimensional functions. Since we are working with
finite sets of input data limited by their physical meaning (the numerical
values characterizing a given task are always from some limited interval), we
can approximate individual exponential functions using the sum of
one-dimensional transition functions of neurons in the first layer and use the
statement \ref{TransitionToSigmoidalActivation} to finally estimate

the size of the network with two hidden layers that approximate the target
function. It is only necessary to use estimates to approximate the exponential
function using standard (e.g., sigmoidal function, ReLU function) transition
functions of neurons in the first hidden layer. There are more such results in
the literature; for example, \cite{EL-ANAN} proposes a procedure for
approximating a one-dimensional function using a sigmoid function with
different limits at infinite points on the real axis. Trivial are then
constructions of approximations of continuous functions using a combination of
ReLU or step functions, where we approximate a continuous function locally on
small intervals.

An idea worth exploring in more detail is to use the constructive proofs of
\ref{lem:approxpolytope} and \ref{thm:fapprox} to design a suitable initial
setting for the weight parameters of a neural network to accelerate the
optimization process of the weight parameters, or to avoid getting stuck in
local minima of the optimized objective function.

Although the practical applicability of the results obtained in this work is
limited, we believe that the presented methodology has the potential to derive
more applicable conclusions for specific, narrower tasks, both for estimating
the size of the required neural networks and for accelerating the relevant
optimization algorithms.

%%%%%%%%%%%%%%%%%%%%%%%%%%%%%%%%%%%%%%%%%%%%%%%%%%%%%%%%%%%%%%%%%%%%%%%%%%%%%%%%%%%%%%%%%%%%%%%%%%%%%%%%%%%%%%%%%%%
%%%%%%% APENDIX %%%%%%%%%%%%%%%%%%%%%%%%%%%%%%%%%%%%%%%%%%%%%%%%%%%%%%%%%%%%%%%%%%%%%%%%%%%%%%%%%%%%%%%%%%%%%%%%%%%
%%%%%%%%%%%%%%%%%%%%%%%%%%%%%%%%%%%%%%%%%%%%%%%%%%%%%%%%%%%%%%%%%%%%%%%%%%%%%%%%%%%%%%%%%%%%%%%%%%%%%%%%%%%%%%%%%%%

\appendix

\section{Necessary Lemmas}
\label{appendix}

The proofs of \ref{lem:approxpolytope} and \ref{thm:fapprox} require more
trivial auxiliary statements, which we present here for clarity. We begin with
several lemmas that enable us to estimate the number of terms required to
approximate polynomials with exponential functions.

\begin{lemma}
\label{lem:cubeexp}
Let $s, c_0 \in \Real{}$, $\sequence{c_i^+}{}\from i=1\to{d}\subset \Real{n}$, $\sequence{c_i^-}{}\from i=1\to{d} \subset \Real{}$, and let
\sequence{\vektor{e}}{i}\from i=1\to{d}, \natur{i}{d} be the standard basis of
\Real{d}\nolinebreak . Define
\begin{equation*}
  \val p\in{\vektor{x}} \defeq c_0 + \sum_{i=1}^{d} c_i^+ e^{s \scalarp{\vektor{e}_i}{\vektor{x}}} +
  \sum_{i=1}^{d} c_i^- e^{s \scalarp{\vektor{e}_i}{\vektor{x}}}.
\end{equation*}
Then, for all $n \in \Natural{}$, $\left(\mapping{p}\right)^n \in \anset{G}_{\exp, h}$, where
\begin{equation*}
    h < 2^d \frac{(n+d)^{n+d}}{n^n d^d} \leq \left( 2e \left( \frac{n}{d} + 1 \right) \right)^d \; .
\end{equation*}
%		\begin{equation*}
%			h < \min\left\{\frac{2^d}{\sqrt{2\pi}}\;\sqrt{\frac{n+1}{n}} \;\frac{(n+1)^{n+1}}{(n-d+1)^{n-d+1}d^d}
%			\quad , \quad
%		\frac{\sqrt{n+2d}\;(n+2d)^{n+2d}}{\sqrt{2\pi}\sqrt{2nd} \;n^n(2d)^{2d}}\right\}\; .
%		\end{equation*}
\end{lemma}
	
\beginproof
 The function \mapping{p} can be written as
\begin{equation*}
	\val p \in{\vektor{x}} = \sum_{i=0}^{2d} c_i e^{\scalarp{\vektor{z}_i}{s\vektor{x}}} \; ,
\end{equation*}
where $\sequence{\vektor{z}}{i}\from 0\to{2d} \defeq \{\vektor{0}, \vektor{e}_1, \dots,
\vektor{e}_d, -\vektor{e}_1,\dots,-\vektor{e}_d\}$ and $\sequence{c}{i}\from 0
\to{2d} \defeq \{c_0,c_1^+,\dots,c_d^+,c_1^-,\dots,c_d^-\}$. 
Denote
\begin{equation*}
    \anset{K}\! \defeq \!\defset{ \!\left(\indexedset{k}{0}{2d}\right) \in
      \left(\Natural{} \cup \left\{0\right\} \right)^{2d+1}}{\sum_{i=1}^{2d}
      k_i = n\!}\; .
\end{equation*}
By the multinomial theorem,
\begin{equation*}
 \!\!\!\! \val p\in{\vektor{x}}^n \!\!=\!\!\!\!\!\!
    \sum_{\left(\indexedset{k}{0}{2d}\right) \in \anset{K}} \frac{n!}{k_0!\cdots k_{2d}!}
   \!\left( \prod_{i=0}^{2d} c_i^{k_i} \!\!\!\right)
   e^{\scalarp{\sum_{i=0}^{2d}k_i\vektor{z}_i}{s\vektor{x}}}  .
\end{equation*}
If we define
$$
\anset{\mathbb Z}_n^d \defeq \defset{\left( \indexedset{m}{1}{d}\right)^T \in
  \Integers{d}}{\sum_{i=1}^d \absvalue{m_i} \leq n}
$$ then
\begin{equation*}
    \sum_{i=0}^{2d}\!\! k_i\vektor{z}_i \!=\! \left(k_1\! -\! k_{d+1},k_2\! - \!
      k_{d+2}, \cdots, k_d\! - \! k_{2d} \right)^T\!\!\! \in \anset{\mathbb Z}_n^d .
\end{equation*}
So the number of terms in $\left(\val p\in{\vektor{x}}\right)^n$ is bounded by
$\setsize{\anset{\mathbb Z}_n^d}$, which by Lemma \ref{ZndSize} is at most
\begin{equation*}
     2^d \frac{(n+d)^{n+d}}{n^n d^d}.
\end{equation*}

\noindent
Further
\begin{equation*}
2^d \frac{\left( n + d \right)^{n+d}}{n^nd^d} = \left( \frac{2}{d}
   \left( n + d \right)\right)^d \left(1 + \frac{d}{n}\right)^n \leq 
\end{equation*}
\begin{equation*}
\leq \left( 2e \left( \frac{n}{d} + 1 \right) \right)^d \; ,
\end{equation*}
where we use the fact that $lim_{\seq n \goto{+\infty}} \left(1+\frac{d}{n} \right)^n = e^d$ and the sequence $\left(1+\frac{d}{n} \right)^n$ is
strictly increasing.

\bigskip
\begin{lemma}
	\label{def:RobbinsBound}
	Let  $n, k \in \mathbb{N}$ such that $n \ge 2$ and $1\leq k\leq n$. Then 
	\begin{equation*}
        {n \choose k} < \frac{n^n}{(n-k)^{n-k}k^k}.
	\end{equation*}
\end{lemma}	
\bigskip
\beginproof
We start with Robbin's version of Stirling's formula \cite{Robbins1955ARemark}
$$
\sqrt{2\pi n}\left(\frac{n}{e}\right)^n e^{\frac{1}{12n+1}} < n! < \sqrt{2\pi
    n}\left(\frac{n}{e}\right)^n e^{\frac{1}{12n}}
$$
for all positive
  integer $n$. Using both bounds we can estimate
$$
{n \choose k} = \frac{n!}{(n-k)!k!} <
$$
$$
< \!\frac{\sqrt{n}\left(\frac{n}{e}\right)^ne^{\frac{1}{12n}}}
{\sqrt{2\pi(n\!-\! k)}\left(\frac{n-k}{e}\right)^{n-k}\!\!e^{\frac{1}{12(n-k)+1}}\!\sqrt{k}\!
  \left(\frac{k}{e}\right)^k\!\!e^{\frac{1}{12k+1}}} =
$$
$$
=\sqrt{\frac{n}{2\pi(n\!-\! k)k}} \;\frac{n^ne^{\left(\frac{1}{12n} -
      \frac{1}{12(n-k)+1} - \frac{1}{12k+1}\right)}}{(n\!-\! k)^{n-k}k^k} \; .
$$
Since for all $k=1,\cdots,n-1$: $n(k-1) \ge (k+1)(k-1)$, we get $(n-k)k \ge n-1$ and therefore $\frac{n}{(n-k)k} \leq
\frac{n}{n-1}$. In addition, $\left(\frac{1}{12n} -
  \frac{1}{12(n-k)+1} - \frac{1}{12k+1}\right) < 0$ for
$k=1,\cdots,n$. So for $n \ge 2$ we get
$$
{n \choose k} \!< \! \sqrt{\frac{n}{2\pi(n\!-\! 1)}}
\frac{n^n}{(n\!-\! k)^{n-k}k^k} \! < \! \frac{n^n}{(n\!-\! k)^{n-k}k^k} \;.
$$
\endproof

\bigskip
\begin{lemma}
	\label{ZndSize}
	Let
        $$\anset{\mathbb Z}_n^k \defeq \defset{(z_1, \dots, z_k)^\top \in
          \mathbb{Z}^k}{\sum_{i=1}^k \absvalue{z_i} \leq n }\;.
        $$ Then 
	\begin{equation*}
        \absvalue{\anset{\mathbb Z}_n^k} < 2^k \frac{(n+k)^{n+k}}{n^n k^k}\; .
	\end{equation*}
\end{lemma}	
\bigskip

\beginproof
Let as denote $\mathbb{N}_{0,k}\defeq (\mathbb{N} \cup \{0\})^k$ 
Using one version of the well-known Balls and Bars Theorem (the number of
configurations of $j$ balls in $k$ bins), we get:
   \begin{equation*}
       \setsize{\left\{\! (z_1, \dots, z_k)^\top\!\!\! \in \mathbb{N}_{0,k} \Big| \sum_{i=1}^k z_i \!=\! j \!\right\}}  = 
        \binom{j+k-1}{k-1}
   \end{equation*}
   and the following property of Pascal's triangle known as the Hockey-stick identity:
   \begin{equation*}
       \sum_{j=0}^n \binom{j+r}{r} = \binom{n+r+1}{r+1},
   \end{equation*}
   we get, by Lemma \ref{def:RobbinsBound}
    \begin{align*}
        \absvalue{\anset{\mathbb Z}_n^k} \le&~ 2^k \absvalue{\left\{ (z_1, \dots, z_k)^\top
                           \!\!\!   \in  \mathbb{N}_{0,k} \Big| \sum_{i=1}^k z_i \leq n \!\right\}} =\\
        =&~ 2^k \!\sum_{j=0}^n \absvalue{\left\{ (z_1, \dots, z_k)^\top \!\!\!
           \in \mathbb{N}_{0,k}\Big| \sum_{i=1}^k z_i \!=\! j \!\right\}}  = \\
      =&~  2^k \!\sum_{j=0}^n \binom{j+k-1}{k-1}   =  2^k \binom{n+k}{k} \lessfrom{\ref{def:RobbinsBound}} \\
       \lessfrom{\ref{def:RobbinsBound}} &~ 2^k \frac{(n+k)^{n+k}}{n^n k^k} \; .
    \end{align*}
\endproof

This work is inspired by the article \cite{borowskydeutsch}, which applies
Bernoulli's inequalities. To clarify the proofs of previous claims, we present
the following lemma, illustrating the principle by which we use Bernoulli's
inequalities in this work.

\bigskip
\begin{lemma}
\label{ApplicationOof_BernoulliInequality}
Let  $\epsilon, \alpha, \beta \in \ooint{0}{1}$, $\alpha < \beta$, $\frac{1}{\alpha} - \frac{1}{\beta} > 1$. Further $a\in \coint{0}{\alpha}$, $b\in \ocint{\beta}{1}$ and
$$
    n \defeq \left\lceil
      \frac{-\log(\epsilon)}{\log\left(2\right)}\right\rceil  \qquad
    \mbox{and} \qquad k \defeq \left\lfloor \frac{1}{\beta} \right\rfloor + 1 \; . 
$$
Then
\begin{equation}
\label{EpsilonGaps}  
(1-a^n)^{k^n} > 1- \epsilon \qquad \mbox{and} \qquad  (1-b^n)^{k^n} < \epsilon \; . 
\end{equation}
\end{lemma}	
\bigskip

\beginproof
The first inequality we prove using the "negative" version of the Bernoulli inequality\footnote{The proof of this version can be done by using the formula for geometric series for $y=(1-x)$ and inequality $m=1+\cdots +1
  \geq 1+y+y^2 + \cdots y^{m-1}$.}
\begin{equation}
\label{BernoulliMinus}
(1-x)^m \geq 1-xm \;,
\end{equation}
where integer $m > 0$ and $0 \leq x \leq 1$. Hence
\begin{equation}
\label{AmKmEquation}
  \left(1-a^m\right)^{k^m} \geq 1 - \left(ka\right)^m > 1-\left(k\alpha\right)^m \; .
\end{equation}

To satisfy the first inequality in \ref{EpsilonGaps} the value of $k$ should meet the inequality $1-\left(k\alpha\right)^m> 1-\epsilon$. Hence
\begin{equation}
 \label{KAlphaRestriction} 
1-\left(k\alpha\right)^m> 1-\epsilon \Leftrightarrow \left(k\alpha\right)^m <
\epsilon  \Leftrightarrow k < \frac{\sqrt[m]{\epsilon}}{\alpha} \; . 
\end{equation}
\noindent
To prove the second inequality we use "positive" version of Bernoulli inequality\footnote{The proof of this version uses a simple inductive argument.}.
\begin{equation}
\label{BernoulliPlus}
  (1+x)^m \geq 1+xm \;, 
\end{equation}
where integer $m > 0$ and $x \geq -1$. So
$$
\left(1-b^m\right)^{k^m} = \frac{k^mb^m}{k^mb^m}\left(1- b^m\right)^{k^m} <
$$
$$
< \frac{1+k^mb^m}{k^mb^m}\left(1-b^m\right)^{k^m}\leqfrom{\ref{BernoulliPlus}} 
$$
$$
\leqfrom{\ref{BernoulliPlus}}  \frac{1}{k^mb^m}\left(1+b^m\right)^{k^m}\left(1-b^m\right)^{k^m} =
$$
$$
= \frac{1}{(kb)^m}\left(1-b^{2m}\right)^{k^m} < \frac{1}{\left(kb\right)^m} < \frac{1}{\left(k\beta\right)^m}  \; .
$$
To satisfy the second inequality in \ref{EpsilonGaps} the value of $k$ should meet the inequality
\begin{equation}
 \label{KBetaRestriction} 
\frac{1}{\left(k\beta\right)^m}  < \epsilon  \Leftrightarrow
\frac{1}{\beta\sqrt[m]{\epsilon}} < k  \; . 
\end{equation}
Putting \ref{KAlphaRestriction} and \ref{KBetaRestriction} together we have
\begin{equation}
\label{BordersForKWithMSquareRoot}
\frac{1}{\beta\sqrt[m]{\epsilon}} < k < \frac{\sqrt[m]{\epsilon}}{\alpha} \; . 
\end{equation}
and if we recall that $\lim_{\seq  m \goto{+\infty}} \sqrt[m]{\epsilon} = 1$
we can estimate the value of $k$ in the range
\begin{equation}
\label{BordersForK}  
\frac{1}{\beta} < k < \frac{1}{\alpha} \; .
\end{equation}
Put $k \defeq \left\lfloor \frac{1}{\beta} \right\rfloor + 1$. Then $k$ is
minimal integral value satisfying inequality $\frac{1}{\beta} < k$, and
because $\frac{1}{\alpha} - \frac{1}{\beta} > 1$, $k$ meets the condition
\ref{BordersForK} and consequently, the condition
\ref{BordersForKWithMSquareRoot} is also true for any values of $m$, so the
conditions \ref{KAlphaRestriction} and \ref{AmKmEquation} are meet.

\vspace{1ex}
\noindent
The first inequality in \ref{BordersForK} follows that $\beta k>1$, so
$\log(k\beta)$ is positive, hence the definition of $k$ follows
\begin{equation}
\label{TwoInsteadOfKBeta}
k\beta \!= \!\left(\left\lfloor\!\frac{1}{\beta}\!\right\rfloor \!+\! 1\!\right)\beta \!<\! 1+\beta \!<\! 2
\Leftrightarrow \frac{1}{\log(2)} \!<\! \frac{1}{\log(k\beta)} \; .
\end{equation}
Now derive value of $m$ such that $\frac{1}{\left(k\beta\right)^m}  < \epsilon $. So
$$
\frac{1}{\left(k\beta\right)^m}  \!< \! \epsilon \!\Leftrightarrow \!
\log\!\left(\!\frac{1}{\left(k\beta\right)^m}\!\right)\!  < \log\!\left(\epsilon\right) \!\Leftrightarrow
\! m \!> \! \frac{-\log(\epsilon)}{\log(k\beta)}  .
$$
and using \ref{TwoInsteadOfKBeta} we have
$$
m > \frac{-\log(\epsilon)}{\log(k\beta)} >  \frac{-\log(\epsilon)}{\log(2)} \; ,
$$
which concludes the proof. 
\endproof

The following two trivial lemmas are used in the proof of statement \ref{thm:fapprox}.

\bigskip
\begin{lemma}
\label{Disjunctive_Sets_Lemma}
Let  $\anset{A}$,  $\anset{B}$,  $\anset{K}$ are given sets and $\anset{A} \subset \anset{B}$. 
Then $\anset{K} \cap \anset{A}$, $\anset{K} \minusset \anset{B}$, $\anset{K}
\cap \left(\anset{B} \minusset \anset{A}\right )$ are mutually disjunctive.
\end{lemma}	

\beginproof
We proove the claim by contradiction.

\noindent
Conditions $x \in \anset{K} \cap \anset{A}$ and  $x \in \anset{K} \minusset \anset{B}$ contradict $\anset{A} \subset \anset{B}$.

\noindent
Conditions $x \in \anset{K} \minusset \anset{B}$ and $x \in \anset{K} \cap \left(\anset{B} \minusset \anset{A}\right )$ follow that $x \notin \anset{B}$ and at the same time $x \in \anset{B}$.

\noindent
Conditions $x \in \anset{K} \cap \anset{A}$ and
$x \in \anset{K} \cap \left(\anset{B} \minusset \anset{A}\right )$ follow that
$x \in \anset{A}$ and at the same time $x \notin \anset{A}$.  \endproof

\begin{lemma}
\label{NumOfIntVectorsInBall}
Let $\anset{B} \defeq
\defset{\vektor{x}\in\Real{d}}{\Euclnorm{\vektor{x}}{}\leq R}$, $R>0$ and $\rho>0$. Then 
\begin{equation*}
\setsize{\rho \Integers{d} \cap \anset{B}} \leq  \frac{\sqrt{\pi}^d\left(R+\rho\frac{\sqrt{d}}{2}\right)^d}{\val \Gamma\in{\frac{d}{2}+1}}\; ,
\end{equation*}
where \mapping{\Gamma} is Euler's gamma function.
\end{lemma}	

\beginproof Denote
$\anset{B}^\prime \defeq
\defset{\vektor{x}\in\Real{d}}{\Euclnorm{\vektor{x}}{}\leq
  R+\rho\frac{\sqrt{d}}{2}}$. For each vector
$\vektor{v} \in \rho \Integers{d} \cap \anset{B}$ define the open cube
$\anset{C}_\vektor{v} \defeq\vektor{v} + \ooint{-\rho\frac{d}{2}}{\rho\frac{d}{2}}^d$. Obviously
$\anset{C}_\vektor{v} \subset \anset{B}^\prime$, so
$\cup_{\vektor{v} \in \rho \Integers{d} \cap
  \anset{B}}\anset{C}_\vektor{v}\subset \anset{B}^\prime$ and
$\setsize{\defset{\anset{C}_\vektor{v}}{\vektor{v} \in \rho \Integers{d} \cap
    \anset{B}}} = \setsize{\rho \Integers{d} \cap \anset{B}}$. In addition,
for $\vektor{v}_1,\vektor{v}_2 \in \rho \Integers{d} \cap \anset{B}$,
$\vektor{v}_1 \neq \vektor{v}_2 $ is
$\anset{C}_{\vektor{v}_1} \cap\anset{C}_{\vektor{v}_2} = \emptyset$. So the
number of different $\anset{C}_\vektor{v}$ is limited by the volume of
$\anset{B}^\prime$, which is
$ \frac{\sqrt{\pi}^d\left(R+\rho\frac{\sqrt{d}}{2}\right)^d}{\val
  \Gamma\in{\frac{d}{2}+1}}\; .  $ \nopagebreak \endproof

The last lemma in this section explains the possibility of using an estimate
of the number of exponential functions needed to approximate a continuous
function on a compact set in \Real{n} to estimate the size of a neural network
using standard sigmoid functions or ReLU functions as neuron transfer
functions.

\begin{lemma}
\label{TransitionToSigmoidalActivation}
Let $\epsilon\in\ooint{0}{1}$, $h,u\in\Natural{}$, $\anset{K} \subset\Real{d}$ be compact, $\mapping{f}\in\Cspace{\anset{K}}$, \funct
\sigma\maps\Real{}\to{\Real{}}. Further let
$$
\EXIST{\left(\vektor{w}_i,c_i\right)\in\karttwoprod{\Real{d}}{\Real{}}, 1\leq i\leq h}
$$
such that 
$$
\ALL{\vektor{x}\in\anset{K}}\HOLDS{\absvalue{\val f\in{\vektor{x}} - \sum_{i=1}^h c_ie^{{}^{\scalarp{\vektor{x}}{\vektor{w}_i}}}} < \frac{\epsilon}{2}}
$$
and for all $\left(\vektor{w}_i,c_i\right)\in\karttwoprod{\Real{d}}{\Real{}}$ exist $\alpha_{j,i},\beta_{j,i},\gamma_{j,i}\in\Real{}$, $1\leq j\leq u$, such that
for all $\vektor{x}\in\anset{K}$ holds
$$
\absvalue{c_ie^{{}^{\scalarp{\vektor{x}}{\vektor{w}_i}}} - 
    \sum_{j=1}^u\alpha_{j,i}\val\sigma\in{\beta_{j,i}\scalarp{\vektor{x}}{\vektor{w}_i}-\gamma_{j,i}}}
  \leq   \frac{\epsilon}{2h}\; .
$$
Then there exists $\mapping{g} \in \anset{G}_{\mapping{\sigma}, hu}$ such that
\begin{equation}
\norm{\mapping{f} - \mapping{g}}{}<\epsilon \; .
\end{equation}
\end{lemma}

\beginproof
$$
\absvalue{\val f\in{\vektor{x}} -  \sum_{i=1}^h  \sum_{j=1}^u\alpha_{j,i}\val\sigma\in{\beta_{j,i}\scalarp{\vektor{x}}{\vektor{w}_i}-\gamma_{j,i}}} \leq
$$
$$  
\leq \absvalue{\val f\in{\vektor{x}} -
  \sum_{i=1}^h c_i e^{{}^{\scalarp{\vektor{x}}{\vektor{w}_i}}}} +
$$
$$
+\absvalue{  \sum_{i=1}^h c_i e^{{}^{\scalarp{\vektor{x}}{\vektor{w}_i}}}
  \!\! - \!\!
   \sum_{i=1}^h \sum_{j=1}^u\alpha_{j,i}\val\sigma\in{\beta_{j,i}\scalarp{\vektor{x}}{\vektor{w}_i}\!-\!\gamma_{j,i}}}\! \leq
$$
$$
\leq \frac{\epsilon}{2}+h\frac{\epsilon}{2h} = \epsilon \; .
$$
\endproof

\bibliographystyle{unsrtnat}
\bibliography{bibliography}  %%% Uncomment this line and comment out the ``thebibliography'' section below to use the external .bib file (using bibtex) .

\end{document}

%% file: NN_common_math_symbols.tex
\ifthenelse{\value{PrintAllSymbols} = 1 }{

    \begin{center}
      {\Large\bf Table of mathematics symbols} \\
        \today
    \end{center}
}{}

\def\printtokens{\futurelet \next\testtoken}        % by Petr Olsak
\def\testtoken{\ifx\next\end \def\detokenize##1{}%
  \else\ifx\next\bgroup \def\detokenize{\char`\{ \ignoretoken}%
  \else\ifx\next\egroup \def\detokenize{\char`\} \ignoretoken}%
  \else\ifx\next\spacetoken \def\detokenize{ \ignoretoken}%
  \else\def\detokenize##1{\string##1\printtokens}\fi\fi\fi\fi
  \detokenize}
\def\ignoretoken{\afterassignment\printtokens\let\next= }
\edef\act{\let\noexpand\spacetoken= \space}\act % \let\spacetoken="space"

\gdef\obsahsymbolu#1#2#3#4#5{  
   \ifthenelse{\value{PrintAllSymbols}=0 \and \equal{1}{#2} }  
              { \noindent \parbox[t]{30mm}{#3} \parbox[t]{120mm}{#5} \\[0.2ex] }
              {}       
   \ifthenelse{\value{PrintAllSymbols}=1 } 
              { \noindent \parbox[t]{25mm}{#3} \parbox[t]{120mm}{\tt \printtokens #3\end} \\ }
              {}       
   \ifthenelse{\value{PrintAllSymbols}=2 }  
              { #1 }
              {}       
}

\gdef\NameOfGroup#1{
   \ifthenelse{\value{PrintAllSymbols}=1 } 
              { \centerline{\rule{\columnwidth}{0.3mm}}
                \centerline{{\normalsize #1}} }
              {}       
}

\newcount\nestedlevel   \nestedlevel=0
\newcount\lastmmode
\gdef\setmathmode{
   \advance\nestedlevel by 1              %  -L=\the\nestedlevel - 
   \ifnum\nestedlevel = 1                   
       \ifmmode \global\lastmmode=1
       \else    \global\lastmmode=0  $
       \fi
   \fi
}

\gdef\restoreoldmode{
   \ifnum\nestedlevel = 1
       \ifnum \lastmmode=0  $ 
       \fi
   \fi
   \advance\nestedlevel by -1             %  -L=\the\nestedlevel -
}

\NameOfGroup{Software}
\obsahsymbolu{\gdef\cmdopt#1#2{ \ifthenelse{\equal{#1}{}}
                                   {\scriptsize\it \hskip -2ex {#2}}   
                                   { {\bf #1}    {\scriptsize\it \hskip -.8ex {#2}}  }  
                              }}
{0}{\cmdopt{-size}{int}}{\cmdopt{-size}{int}}{}
\obsahsymbolu{\gdef\fname#1{ \underline{\sc #1}}} 
{0}{\fname{manhattan.gif}}{}{}
\NameOfGroup{Miscelany}
\obsahsymbolu{

}
{0}%{\begin{fcase}{lcl} \fcaseitem{1}{}{a} \fcaseitem{2}{pro}{b} \fcaseitem{3}{}{c} \end{fcase}}{}{}
{}{}{}
\obsahsymbolu{\gdef\defeq{\setmathmode \stackrel{{\mbox{{\tiny \resizebox{2ex}{!}{def}}}}}{=} \restoreoldmode}}
{1}{\defeq}{}{definition of newly introducec object, set or number}
\obsahsymbolu{\gdef\implyfrom#1{\setmathmode \stackrel{{\mbox{\tiny #1}}}{\Rightarrow} \restoreoldmode}}
{1}{\implyfrom{2}}{}{implication from equation, etc.}
\obsahsymbolu{\gdef\eqfrom#1{\setmathmode \stackrel{{\mbox{\tiny #1}}}{=} \restoreoldmode}}
{1}{\eqfrom{4}}{}{equality follows from 4}
\obsahsymbolu{\gdef\neqfrom#1{\setmathmode \stackrel{{\mbox{\tiny #1}}}{\neq} \restoreoldmode}}
{1}{\neqfrom{5}}{}{nonequality follows from 5}
\obsahsymbolu{\gdef\leqfrom#1{\setmathmode \stackrel{{\mbox{\tiny #1}}}{\leq} \restoreoldmode}}
{1}{\leqfrom{xx}}{}{follows from xx}
\obsahsymbolu{\gdef\geqfrom#1{\setmathmode \stackrel{{\mbox{\tiny #1}}}{\geq} \restoreoldmode}}
{1}{\geqfrom{rs}}{}{follows from rs}
\obsahsymbolu{\gdef\lessfrom#1{\setmathmode \stackrel{{\mbox{\tiny #1}}}{<} \restoreoldmode}}
{1}{\lessfrom{xx}}{}{less than follows from xx}
\obsahsymbolu{\gdef\greatfrom#1{\setmathmode \stackrel{{\mbox{\tiny #1}}}{>} \restoreoldmode}}
{1}{\greatfrom{xx}}{}{great than follows from xx}
\obsahsymbolu{
\gdef\initstep#1{\setmathmode \newline \noindent #1 : \newline \noindent \restoreoldmode}}
{0}{\initstep{k=1}}{}{}
\obsahsymbolu{\gdef\induction#1#2{\setmathmode \\ \noindent #1 \mapsto #2 : \\ \\  \noindent \restoreoldmode}}
{0}{\induction{k}{k+1}}{}{}
\obsahsymbolu{
\gdef\poly#1{{\setmathmode  polytime \left( #1 \right) \restoreoldmode}}
}{0}{\poly{n}}{}{}
\obsahsymbolu{\gdef\INSTANCE#1#2{ \\[2mm] \noindent {\bf INSTANCE:}{\sc #1} \\ \noindent {#2}}}
{0}{\INSTANCE{AAA}{Big thing.}}{}{}
\obsahsymbolu{\gdef\QUESTION#1{ \\[2mm]\noindent {\bf PROBLEM:} \\  \noindent {#1}\\}}
{0}{\QUESTION{What is it?}}{}{}
\obsahsymbolu{\gdef\ALGORITHM#1#2{ \begin{center} ALGORITHM  \newline {\bf {#1}} \end{center} {\sl {#2}}}}
{0}{\ALGORITHM{Long life:}{Do not smoke!}}{}{}
\obsahsymbolu{\gdef\ALL#1{\setmathmode \left( \forall #1 \right)\restoreoldmode}}
{0}{\ALL{\vektor{x} \in \anset{A}}}{}{}
\obsahsymbolu{\gdef\EXIST#1{\setmathmode \left( \exists #1 \right)\restoreoldmode}}
{0}{\EXIST{\mapping{F}}}{}{}
\obsahsymbolu{\gdef\HOLDS#1{\setmathmode \left( #1 \right)\restoreoldmode}}
{0}{\HOLDS{\alpha < \beta }}{}{}
\NameOfGroup{DEFINITIONS and TEXTS}
\obsahsymbolu{\gdef\vsuvka#1{ \footnote{#1} }}
{0}{\vsuvka{text vsuvky}}{}{}
\obsahsymbolu{\gdef\contextdef#1{ \ifnum\finalversion = 0 {\Large\sf DEF: #1} \fi }}
{0}{\contextdef{DEF cosi}}{}{}
\obsahsymbolu{\gdef\provedin#1#2{ \noindent \rule[2pt]{1.5mm}{1.5mm} Proved in: {\sl #1} \cite{#2}}}
{0}{\provedin{Koran}}{}{}
\obsahsymbolu{\gdef\theodef#1{  {\bf #1 (Definition)} }}
{0}{\theodef{cosi}}{}{}
\obsahsymbolu{\gdef\nn#1#2{ {\sc #1} \index{#2} }}
{0}{\nn{novy pojem}{novy pojem}}{}{}
\obsahsymbolu{\gdef\english#1{}}
{0}{\english{mouse}}{}{}
\obsahsymbolu{\gdef\valueof#1{{\setmathmode {\tt #1} \restoreoldmode}}}
{0}{\valueof{Z}}{}{}
\obsahsymbolu{
\gdef\userlabel#1{
 \label{#1}
 \ifnum\finalversion = 0
   \setmathmode \stackrel{{\bf LABEL}}{{\small\bf #1}} \restoreoldmode
 \fi
}}
{0}{\userlabel{Odkaz}}{}{}
\obsahsymbolu{\gdef\privatenote#1{ \ifnum\finalversion = 0 {\Large\sc PRIVATE: #1} \\ \fi}}
{0}{\privatenote{!!!}}{}{}
\obsahsymbolu{\newtheorem{definition}{{\colorbox{white!40}{Definition}}}[section]}
{0}{}{}{}
\obsahsymbolu{\newtheorem{theorem}{\colorbox{white!40}{Theorem}}[section]}
{0}{}{}{}
\obsahsymbolu{\newtheorem{lemma}[theorem]{\colorbox{white!40}{Lemma}}}
{0}{}{}{}
\obsahsymbolu{}
{0}{}{}{}
\obsahsymbolu{}
{0}{}{}{}
\obsahsymbolu{}
{0}{}{}{}
\obsahsymbolu{}
{0}{}{}{}
\obsahsymbolu{

}{0}{}{}{}
\NameOfGroup{PROOFS}
\obsahsymbolu{\gdef\beginproof{ \noindent \colorbox{white!15}{\rule{1.5mm}{1.5mm} {\it Proof:}} \\ }}
{0}{\beginproof}{}{}
\obsahsymbolu{\gdef\proofpart#1{ \noindent \rule{1mm}{1mm} add #1) \\ }}
{0}{\proofpart{a}}{}{}
\obsahsymbolu{\gdef\endproof{ \rule{0mm}{0mm} \\ \rule{0mm}{0mm} \hfill \colorbox{white!15}{\rule[0.5ex]{2.5mm}{0.3mm}   {\it q.e.d.} \rule[0.5ex]{2.5mm}{0.3mm}} \\ }}
{0}{\endproof}{}{}
\NameOfGroup{SETS}
\obsahsymbolu{\gdef\natur#1#2{{\setmathmode  #1\! \in\! \{1,\dots,#2\} \restoreoldmode}}}
{0}{\natur{\alpha}{n}}{natur\{alpha\}\{n\}}{}
\obsahsymbolu{\gdef\naturzero#1#2{{\setmathmode #1\! \in\! \{0,\dots,#2\} \restoreoldmode}}}
{0}{\naturzero{\alpha}{2^n}}{}{}
\obsahsymbolu{\gdef\naturdef#1#2{{\setmathmode #1\! = \{1,\dots,#2\} \restoreoldmode}}}
 {1}{\naturdef{K}{m}}{}{set of natural numbers between $1$ and $m$}
\obsahsymbolu{\gdef\naturset#1#2{{\setmathmode #1\! \subset \{1,\dots,#2\} \restoreoldmode}}}
 {0}{\naturset{\omega}{m}}{}{}
\obsahsymbolu{\gdef\Bool#1{{\setmathmode \{0,1\}^{#1} \restoreoldmode}}}
 {0}{\Bool{N}}{}{}
\obsahsymbolu{\gdef\Pmone#1{{\setmathmode \{-1,+1\}^{#1} \restoreoldmode}}}
 {0}{\Pmone{k}}{}{}
\obsahsymbolu{\gdef\Ball#1#2{{\setmathmode B^{#1}_{#2} \restoreoldmode}}}
 {1}{\Ball{N}{\alpha}}{}{ball in \Real{N} space centered at initial point and with radius $\alpha$}
\obsahsymbolu{\gdef\Real#1{{\setmathmode \Re^{#1} \restoreoldmode}}}
 {1}{\Real{N}}{}{real $N$-dimensional space}
\obsahsymbolu{\gdef\Realp#1{{\setmathmode \Re_+^{#1} \restoreoldmode}}}
 {0}{\Realp{N}}{}{}
\obsahsymbolu{\gdef\Realpz#1{{\setmathmode \Re_{0+}^{#1} \restoreoldmode}}}
 {0}{\Realpz{N}}{}{}
\obsahsymbolu{\gdef\Natural#1{{\setmathmode \mathbb{N}^{#1} \restoreoldmode}}}
 {0}{\Natural{2}}{}{positive integers}
\obsahsymbolu{\gdef\Naturalzero#1{{\setmathmode \mathbb{N}_0^{#1} \restoreoldmode}}}
 {0}{\Naturalzero{2}}{}{nonnegative positive integers}
\obsahsymbolu{\gdef\Integers#1{{\setmathmode \mathbb{Z}^{#1} \restoreoldmode}}}
 {0}{\Integers{2}}{}{all integers}
\obsahsymbolu{\gdef\kartpr#1#2{{\setmathmode {#1}^{#2} \restoreoldmode}}}
 {1}{\kartpr{\Omega}{m}}{}{$m$-tuple Cartesian product of set $\Omega$}
\obsahsymbolu{\gdef\kartpoint#1{{\setmathmode \stackrel{\smile}{#1} \restoreoldmode}}}
 {1}{\kartpoint{z}}{}{element in Cartesian (multiple) product of specified set}
\obsahsymbolu{\gdef\anset#1{{\setmathmode \bar{#1} \restoreoldmode}}}
 {1}{\anset{X}}{anset\{X\}}
{general (sub)set of a given set}
\obsahsymbolu{\gdef\colset#1{{\setmathmode {\sf #1} \restoreoldmode}}}
 {1}{\colset{X}}{colset\{X\}}
{system of subsets of a given set}
\obsahsymbolu{\gdef\colcolset#1{{\setmathmode {\cal #1} \restoreoldmode}}}
 {1}{\colcolset{X}}{}{set of systems of sets over a given set}
\obsahsymbolu{\gdef\potence#1{{\setmathmode 2^{#1} \restoreoldmode}}}
 {1}{\potence{\anset{X}}}{}{potential set of \anset{X}, $\potence{\anset{X}}\defeq\defset{\anset{Z}}{\anset{Z} \subset \anset{X}}$}
\obsahsymbolu{\gdef\ooint#1#2{{\setmathmode  \left( #1 , #2 \right)
    \restoreoldmode}}}
 {0}{\ooint{a}{b}}{}{open interval \ooint{a}{b}}
\obsahsymbolu{\gdef\ccint#1#2{{\setmathmode \left\langle #1 ,#2 \right\rangle \restoreoldmode}}}
 {0}{\ccint{a}{b}}{}{closet interval \ccint{a}{b}}
\obsahsymbolu{\gdef\ocint#1#2{{\setmathmode  \left( #1 , #2 \right\rangle \restoreoldmode}}}
 {0}{\ocint{a}{b}}{}{open-closed interval \ocint{a}{b}}
\obsahsymbolu{\gdef\coint#1#2{{\setmathmode  \left\langle #1 , #2 \right) \restoreoldmode}}}
 {0}{\coint{a}{b}}{}{closed-open interval \coint{a}{b}}
\obsahsymbolu{\gdef\linhull#1{{\setmathmode \left[#1\right]_{\lambda} \restoreoldmode}}}
 {1}{\linhull{\Omega}}{}{linear hull of set $\Omega$ of vectors from
 linear vector space (=set of all linear combination of finite set of vectors from $\Omega$)}
\obsahsymbolu{\gdef\ortcompl#1{{\setmathmode {#1}^{\bot} \restoreoldmode}}}
 {1}{\ortcompl{\anset{B}}}{}{orthogonal complement of the set \anset{B}}
\obsahsymbolu{\gdef\conhull#1{{\setmathmode [#1]_{\kappa} \restoreoldmode}}}
 {1}{\conhull{\anset{B}}}{}{convex hull of set \anset{B}}
\obsahsymbolu{\gdef\vektspace#1#2#3{{\setmathmode {\cal #1}_{#2}^{#3} \restoreoldmode}}}
 {1}{\vektspace{H}{n}{*}}{}{vektor space (with specified parameters)}
\obsahsymbolu{\gdef\Cspace#1{{\setmathmode C_{#1} \restoreoldmode}}}
 {1}{\Cspace{\anset{A}}}{}{set of continuous functions defined on a given set \anset{A}}
\obsahsymbolu{\gdef\CLASS#1#2#3{{\setmathmode {\bf #1}_{#2}^{#3} \restoreoldmode}}}
 {1}{\CLASS{NP}{k}{\alpha}}{}{set of objest with predefined
 properties, generally set of functions, mappings, etc., context dependent}
\obsahsymbolu{\gdef\proces#1{{\setmathmode {\mbox{{\sc #1}}}^* \restoreoldmode}}}
 {1}{\proces{B}}{}{general learning algorithm}
\obsahsymbolu{\gdef\indexedset#1#2#3{{\setmathmode {#1}_{#2}, \ldots ,{#1}_{#3} \restoreoldmode}}}
 {0}{\indexedset{\colset{A}}{i}{5}}{}{}
\obsahsymbolu{\gdef\defset#1#2{{\setmathmode \left\{ {#1} \left| {#2} \right. \right\} \restoreoldmode}}}
 {0}{\defset{\omega\in\Real{}}{\omega >0}}{}{}
\obsahsymbolu{\gdef\compl#1{{\setmathmode \restoreoldmode}}}
 {0}{\compl{\anset{A}}}{}{}
\obsahsymbolu{\gdef\symdiff#1#2{{\setmathmode {#1} \bigtriangleup {#2} \restoreoldmode}}}
 {1}{\symdiff{\anset{A}}{\anset{B}}}{}{symetric difference of two sets \anset{A} and \anset{B}}
\obsahsymbolu{\gdef\minusset{{\setmathmode  \stackrel{.}{-} \restoreoldmode}}}
 {1}{$\anset{A} \minusset \anset{B}$}{}{difference of sets \anset{A} and \anset{B}}
\NameOfGroup{SEQUENCES}
\obsahsymbolu{\gdef\sequence#1#2\from#3\to#4{{\setmathmode \{ {#1}_{#2} \}_{#3}^{#4} \restoreoldmode}}}
 {0}{\sequence{a}{i}\from n \to{k}}{}{}
\obsahsymbolu{\gdef\zeroorder#1{{\setmathmode o\left( #1 \right) \restoreoldmode}}}
 {1}{\zeroorder{\mapping{f}}}{}{$\mapping{g}=\zeroorder{\mapping{f}}$
 means that absolute value of function \mapping{g} is on assymptotic neighbourhood of
 zero majorized by function \mapping{f}}
\obsahsymbolu{\gdef\inforder#1{{\setmathmode O\left( #1 \right) \restoreoldmode}}}
 {1}{\inforder{\mapping{f}}}{}{$\mapping{g}=\inforder{\mapping{f}}$
 means that absolute value of function \mapping{g} is on assymptotic
 neighbourhood of infinity majorised be values of \mapping{f}}
\obsahsymbolu{\gdef\omegaorder#1{{\setmathmode \Omega\left( #1 \right)
    \restoreoldmode}}}
 {0}{\omegaorder{\eta}}{}{}
\obsahsymbolu{\gdef\thetaorder#1{{\setmathmode \Theta\left( #1 \right)
    \restoreoldmode}}}
 {0}{\thetaorder{delta}}{}{}
\obsahsymbolu{\gdef\sameorder#1{{\setmathmode \approx\left( #1
    \right)\restoreoldmode}}}
 {0}{\sameorder{\delta}}{}{}
\obsahsymbolu{\gdef\seq#1\goto#2{{\setmathmode {#1} \rightarrow {#2} \restoreoldmode}}}
 {0}{\seq m \goto {\infty}}{}{}
\NameOfGroup{MATRICES and VECTORS}
\obsahsymbolu{\gdef\matr#1{{\setmathmode \mbox{\boldmath$#1$} \restoreoldmode}}}
 {1}{\matr{A}}{}{matrix of real numbers, colums or rows will be
 denoted by notion \nn{lines of matrix}{lines of matrix} }
\obsahsymbolu{\gdef\vektor#1{{\setmathmode \vec{\mbox{\boldmath$#1$}} \restoreoldmode}}}
 {1}{\vektor{x}}{}
 {vector from a given vector space (all vectors will be considerd as columns)}
\obsahsymbolu{\gdef\vindice#1#2{{\setmathmode \vec{\mbox{\boldmath$#1$}}_#2 \restoreoldmode}}}
 {1}{\vindice{x}{i}}{}
 {i-th indice of the vector)}
\obsahsymbolu{\gdef\expmatr#1#2{{\setmathmode \matr{#1}^{\left( #2 \right)} \restoreoldmode}}}
 {1}{\expmatr{A}{n}}{}{square matrix of order $2^n$}
\obsahsymbolu{\gdef\expvektor#1#2{{\setmathmode \vektor{#1}^{\left( #2 \right)} \restoreoldmode}}}
 {1}{\expvektor{x}{n}}{}{vector of dimension $2^n$}
\obsahsymbolu{\gdef\matri#1#2{{\setmathmode
   {}^{#2}\!\matr{#1} %  #1 name,  #2 index
\restoreoldmode}}}
 {0}{\matri{A}{i}}{}
{}
\obsahsymbolu{\gdef\matsys#1#2{{\setmathmode
    \matri{#1}{1},\dots,\matri{#1}{#2}
\restoreoldmode}}}
 {0}{\matsys{A}{N}}{}
{}
\obsahsymbolu{\gdef\matsysi#1#2#3{{\setmathmode
    {\matri{#1}{1}}_{#3},\dots,{\matri{#1}{#2}}_{#3}
\restoreoldmode}}}
 {0}{\matsysi{A}{N}{i}}{}
{}
\obsahsymbolu{\gdef\mattwo#1#2#3#4{{\setmathmode \left( {\hfill {#1} \atop \hfill {#3}}
  {\hfill {#2} \atop \hfill {#4}} \right)\restoreoldmode}}}
 {0}{\mattwo{3}{-4}{\frac{1}{2}}{\pi}}{}{}
\obsahsymbolu{\gdef\dettwo#1#2#3#4{{\setmathmode % det of order 2
  \left| { \hfill {#1} \atop \hfill {#3}}
         { \hfill {#2} \atop \hfill {#4}}
  \right|
\restoreoldmode}}}
 {0}{\dettwo{3}{-4}{\frac{1}{2}}{\pi}}{}{}
\obsahsymbolu{\gdef\matthree#1#2#3#4#5#6#7#8#9{{\setmathmode  % by rows
   {\small \left(
    \begin{array}{rrr}
       #1 & #2 & #3 \\ #4 & #5 & #6 \\ #7 & #8 & #9
    \end{array}  \right) }
\restoreoldmode}}}
 {0}{\matthree{1}{2}{3}{-4}{5}{7}{12}{0}{\alpha}}{}{}
\obsahsymbolu{\gdef\detthree#1#2#3#4#5#6#7#8#9{{\setmathmode  % by rows
   {\small
    \begin{array}{|rrr|}
       #1 & #2 & #3 \\ #4 & #5 & #6 \\ #7 & #8 & #9
    \end{array}   }
\restoreoldmode}}}
 {0}{\detthree{1}{2}{3}{-4}{5}{7}{12}{0}{\alpha}}{}{}
\obsahsymbolu{\gdef\dimbydim#1#2{{\setmathmode #1\!\times \!#2 \restoreoldmode}}}
 {0}{\dimbydim{n}{m}}{}{}
\obsahsymbolu{\gdef\VektorPar#1{ \setmathmode {\tilde{\expvektor{p}{#1}}}
    \restoreoldmode }}
 {0}{\VektorPar{n}}{}{}
\obsahsymbolu{\gdef\SdruzMat#1{ \setmathmode \matr{M}_{#1} \restoreoldmode }}
 {0}{\SdruzMat{y}}{}{}
\obsahsymbolu{\gdef\SylvestrMat#1{ \setmathmode \expmatr{B}{#1} \restoreoldmode }}
 {0}{\SylvestrMat{n}}{}{}
\NameOfGroup{OPERATORS}
\obsahsymbolu{\gdef\scalarp#1#2{{\setmathmode \left\langle {#1} \left| {#2} \right. \right\rangle \restoreoldmode}}}
 {1}{\scalarp{\vektor{x}}{\vektor{y}}}{}{scalar product of vectors,  $\scalarp{\vektor{x}}{\vektor{y}}= \sum_{i=1}^n \vektor{x}_i \vektor{y}_i$}
\obsahsymbolu{\gdef\tensorp#1#2{{\setmathmode {#1} \odot {#2} \restoreoldmode}}}
 {1}{\tensorp{\vektor{x}}{\vektor{y}}}{}{tensor product of vectors or matrices}
\obsahsymbolu{\gdef\karttwoprod#1#2{{\setmathmode {#1} \times {#2} \restoreoldmode}}}
 {1}{\karttwoprod{\anset{X}}{\anset{M}}}{}{cartesian product of two sets}
\obsahsymbolu{\gdef\corelation#1#2{{\setmathmode
  \valueof{cor}\left( {#1}, {#2} \right)
\restoreoldmode}}}
 {0}{\corelation{\matr{A}}{\matr{B}}}{}
{}
\obsahsymbolu{\gdef\boolplus{{\setmathmode \oplus \restoreoldmode}}}
 {0}{\boolplus}{}
{}
\obsahsymbolu{\gdef\boolmult{{\setmathmode \otimes \restoreoldmode}}}
 {0}{\boolmult}{}
{}
\obsahsymbolu{\gdef\logAND{{\setmathmode \stackrel{\wedge}{\mbox{{\scriptsize AND}}} \restoreoldmode}}}
 {1}{\logAND}{}{binary function AND}
\obsahsymbolu{\gdef\logOR {{\setmathmode \stackrel{\vee}{\mbox{{\scriptsize OR}}} \restoreoldmode}}}
 {1}{\logOR}{}{binary function OR}
\obsahsymbolu{\gdef\logXOR{{\setmathmode  \stackrel{\sqcup}{\mbox{{\scriptsize XOR}}} \restoreoldmode}}}
 {1}{\logXOR}{}{binary function XOR}
\obsahsymbolu{\gdef\logNOT{{\setmathmode \stackrel{\neg}{\mbox{{\scriptsize NOT}}} \restoreoldmode}}}
 {1}{\logNOT}{}{unary function NOT}
\obsahsymbolu{\gdef\bellowpart#1{{\setmathmode \left\lfloor #1 \right\rfloor \restoreoldmode}}}
 {1}{\bellowpart{M}}{}{greatest whole number less than $M$ (=whole bellow part)}
\obsahsymbolu{\gdef\upperpart#1{{\setmathmode \left\lceil #1 \right\rceil \restoreoldmode}}}
 {1}{\upperpart{M}}{}{smallest whole number greater than $M$ (=whole upper part)}
\obsahsymbolu{\gdef\vpm#1{{\setmathmode vpm\left(#1\right) \restoreoldmode}}}
 {1}{\vpm{j}}{}{\vpm{j} is vector from \Pmone{n},
  (resp. \Bool{n}, by context), whose coordinates correspond with
 binary inscription of natural number $j$ (coordinates of vector \vpm{j}
  correspond to pozitions of $1$ in binary expansion of $j$, are
 equal to $1$).}
\obsahsymbolu{\gdef\binvekof#1{{\setmathmode vbin\left(#1\right) \restoreoldmode}}}
 {1}{\binvekof{j}}{}{\binvekof{j} is vector from {0,1},
  whose coordinates correspond with
 binary inscription of natural number $j$ (coordinates of vector \binvekof{j}
  correspond to pozitions of $1$ in binary expansion of $j$, are
 equal to $1$).}
\obsahsymbolu{\gdef\intnumof#1{{\setmathmode nint\left(#1\right) \restoreoldmode}}}
 {1}{\intnumof{\vektor{v}}}{}{whole number, derived from vector $\vektor{v}\in
  \Pmone{n}$, (resp. $\vektor{v}\in \Bool{n}$, by context), where
  ones components of vector \intnumof{\vektor{v}}
  correspond to $1$ in binary expansion of number
 \intnumof{\vektor{v}}, the rest components correspond to position of
 $0$ in binary expansion of  \intnumof{\vektor{v}}}
\obsahsymbolu{\gdef\BinCoef#1#2{{\setmathmode {#1 \choose #2} \restoreoldmode}}}
 {1}{\BinCoef{a}{i}}{}{binomical coefficien}
\obsahsymbolu{\gdef\operator#1#2#3{{\setmathmode
   {\sl #1}_{#2}^{#3}
\restoreoldmode}}}
 {0}{\operator{CONV}{n}{L}}{}
{}
\NameOfGroup{MAPPINGS and FUNCTIONS}
\obsahsymbolu{\gdef\mapping#1{{\setmathmode \widetilde{#1} \restoreoldmode}}}
 {1}{\mapping{f}}{}{general mapping between two sets}
\obsahsymbolu{\gdef\konvoluce#1#2{{\setmathmode \left(\mapping{#1} * \mapping{#2}\right) \restoreoldmode}}}
 {1}{\konvoluce{f}{g}}{}{convolution of functions \mapping{f} a \mapping{g}}
\obsahsymbolu{\gdef\val#1\in#2{{\setmathmode \mapping{#1}\left( {#2}
      \right) \restoreoldmode}}}
 {1}{\val A \in {b}}{}{value of function (mapping) \mapping{A} at point $b$}
\obsahsymbolu{\gdef\blankval#1\in#2{{\setmathmode #1\left( {#2} \right) \restoreoldmode}}}
 {}{\blankval G \in {b}}{}{value of expression G in argument $b$}
\obsahsymbolu{\gdef\funct#1\maps#2\to#3{{\setmathmode
   \mapping{#1} : #2 \rightarrow #3
\restoreoldmode}}}
 {0}{\funct\Gamma\maps\Bool{N}\to{\Omega^2}}{}{}
\obsahsymbolu{\gdef\I#1\h#2\f#3\d#4\x#5{{\setmathmode
   \int_{#1}^{#2} {#3 {#4}\left({#5}\right)}
\restoreoldmode}}}
 {0}{\I a \h b \f \alpha \d \Gamma \x{z}}{}{}
\obsahsymbolu{\gdef\Fourierof#1{{\setmathmode \widehat{ \mapping{#1} } \restoreoldmode}}}
 {0}{\Fourierof{G}}{}
{}
\obsahsymbolu{\gdef\natlog#1{{\setmathmode \ln \left( #1 \right) \restoreoldmode}}}
 {1}{\natlog{x}}{}{natural logarithm of x}
\obsahsymbolu{\gdef\binlog#1{{\setmathmode \log_2 \left( #1 \right) \restoreoldmode}}}
 {1}{\binlog{x}}{}{logarithm of radix 2}
\obsahsymbolu{\gdef\modulo#1#2#3{{\setmathmode #1 \equiv #2 \quad mod \quad #3 \restoreoldmode}}}
 {1}{\modulo{x}{y}{r}}{}{\modulo{x}{y}{r} means that there exist whole
 number $k$ such that $x=yk+r$, where $x$, $y$, and $r$ are whole numbers}
\obsahsymbolu{\gdef\Prob#1#2{{\setmathmode Prob_{#1}\left( #2 \right) \restoreoldmode}}}
 {1}{\Prob{\mapping{\Pi}}{\anset{A}}}{}{probability of set \anset{A}
 under probability distribution \mapping{\Pi}}
\obsahsymbolu{\gdef\setsize#1{{\setmathmode \left| {#1} \right| \restoreoldmode}}}
 {1}{\setsize{\anset{V}}}{}{cardinality of set \anset{V}}
\obsahsymbolu{\gdef\absvalue#1{{\setmathmode \left| {#1} \right| \restoreoldmode}}}
 {1}{\absvalue{\delta}}{}{absolute value of number $\delta$}
\NameOfGroup{NORMS}
\obsahsymbolu{\gdef\maxnorm#1{{\setmathmode  \left\| #1 \right\|_{max} \restoreoldmode}}}
 {1}{\maxnorm{\vektor{z}}}{}{maximum norm of vector}
\obsahsymbolu{\gdef\sumnorm#1{{\setmathmode  \left\| #1 \right\|_{\Sigma}
    \restoreoldmode}}}
 {0}{\sumnorm{\vektor{z}}}{}
{}
\obsahsymbolu{\gdef\Euclnorm#1{{\setmathmode \left\| #1 \right\|_{E} \restoreoldmode}}}
 {1}{\Euclnorm{\vektor{z}}}{}{Eucleidian norm of vector}
\obsahsymbolu{\gdef\norm#1#2{{\setmathmode   \left\| #1 \right\|_{#2} \restoreoldmode}}}
 {1}{\norm{\vektor{z}}{\vektspace{L}{2}{\anset{K}}}}{}{specified norm of vector}
\NameOfGroup{GEOMETRY}
\obsahsymbolu{\gdef\volume#1{{\setmathmode  \valueof{Vol}\left( #1 \right)
    \restoreoldmode}}}
 {0}{\volume{\Ball{m}{1}}}{}
{}
\obsahsymbolu{\gdef\diameter#1{{\setmathmode  \valueof{Diam}\left( #1 \right)
    \restoreoldmode}}}
 {0}{\diameter{\Ball{m}{1}}}{}
{}
\obsahsymbolu{\gdef\project#1#2{{\setmathmode  \left.{#1}\right|_{#2} \restoreoldmode}}}
 {1}{\project{\anset{A}}{\vektspace{P}{}{}}}{}
{orthogonal projection of set \anset{A} into subspace \vektspace{P}{}{}}
\NameOfGroup{STATISTICS}
\obsahsymbolu{\gdef\meanvalue#1#2{{\setmathmode  \mathbf{E}_{#1}\left[#2\right] \restoreoldmode}}}
 {1}{\meanvalue{x}{\mapping{P}}}{}{mean value of random variable}
\obsahsymbolu{\gdef\undercondition#1{{\setmathmode  \left|#1\right. \restoreoldmode}}}
 {1}{\undercondition{\mapping{P}}}{}{under condition}

%% file: aproximation-on-x-neigbourghoud.pdf_t
\begin{picture}(0,0)%
\includegraphics{aproximation-on-x-neigbourghoud.pdf}%
\end{picture}%
\setlength{\unitlength}{4144sp}%
\begingroup\makeatletter\ifx\SetFigFont\undefined%
\gdef\SetFigFont#1#2#3#4#5{%
  \reset@font\fontsize{#1}{#2pt}%
  \fontfamily{#3}\fontseries{#4}\fontshape{#5}%
  \selectfont}%
\fi\endgroup%
\begin{picture}(5655,1791)(256,-3955)
\put(5896,-2446){\makebox(0,0)[lb]{\smash{{\SetFigFont{12}{14.4}{\rmdefault}{\mddefault}{\updefault}{\color[rgb]{0,0,0}$1$}%
}}}}
\put(5896,-3571){\makebox(0,0)[lb]{\smash{{\SetFigFont{12}{14.4}{\rmdefault}{\mddefault}{\updefault}{\color[rgb]{0,0,0}$0$}%
}}}}
\put(271,-3166){\makebox(0,0)[lb]{\smash{{\SetFigFont{12}{14.4}{\rmdefault}{\mddefault}{\updefault}{\color[rgb]{0,0,0}$\frac{\gamma}{2}$}%
}}}}
\put(271,-2806){\makebox(0,0)[lb]{\smash{{\SetFigFont{12}{14.4}{\rmdefault}{\mddefault}{\updefault}{\color[rgb]{0,0,0}$\gamma$}%
}}}}
\put(5131,-3841){\makebox(0,0)[lb]{\smash{{\SetFigFont{12}{14.4}{\rmdefault}{\mddefault}{\updefault}{\color[rgb]{0,0,0}$x_0^i+\omega r$}%
}}}}
\put(801,-3886){\makebox(0,0)[lb]{\smash{{\SetFigFont{12}{14.4}{\rmdefault}{\mddefault}{\updefault}{\color[rgb]{0,0,0}$x_0^i-\omega r$}%
}}}}
\put(2986,-3751){\makebox(0,0)[lb]{\smash{{\SetFigFont{12}{14.4}{\rmdefault}{\mddefault}{\updefault}{\color[rgb]{0,0,0}$x_0^i$}%
}}}}
\put(1531,-3886){\makebox(0,0)[lb]{\smash{{\SetFigFont{12}{14.4}{\rmdefault}{\mddefault}{\updefault}{\color[rgb]{0,0,0}$x_0^i-r$}%
}}}}
\put(4446,-3841){\makebox(0,0)[lb]{\smash{{\SetFigFont{12}{14.4}{\rmdefault}{\mddefault}{\updefault}{\color[rgb]{0,0,0}$x_0^i+r$}%
}}}}
\end{picture}%